\def\year{2022}\relax
\documentclass[letterpaper]{article} 
\usepackage{aaai22}  
\usepackage{times}  
\usepackage{helvet}  
\usepackage{courier}  
\usepackage[hyphens]{url}  
\usepackage{graphicx} 
\usepackage{natbib}  
\usepackage{caption} 
\DeclareCaptionStyle{ruled}{labelfont=normalfont,labelsep=colon,strut=off} 
\usepackage{algorithm}
\usepackage{algorithmic}
\usepackage[caption=false, font=footnotesize, position=top]{subfig}
\usepackage{amssymb}
\usepackage[flushleft]{threeparttable}
\usepackage{enumitem}
\usepackage{multirow}
\usepackage{multicol}
\usepackage{booktabs}

\usepackage{newfloat}
\usepackage{listings}
\floatstyle{ruled}
\newfloat{listing}{tb}{lst}{}
\floatname{listing}{Listing}
\title{PRMI: A Dataset of Minirhizotron Images for Diverse Plant Root Study}
\author{
    Weihuang Xu,\textsuperscript{\rm 1}
    Guohao Yu,\textsuperscript{\rm 1}
    Yiming Cui,\textsuperscript{\rm 1}
    Romain Gloaguen,\textsuperscript{\rm 2}
    Alina Zare,\textsuperscript{\rm 1} 
    Jason Bonnette,\textsuperscript{\rm 3}
    Joel Reyes-Cabrera,\textsuperscript{\rm 4}
    Ashish Rajurkar,\textsuperscript{\rm 4}
    Diane Rowland,\textsuperscript{\rm 5}
    Roser Matamala,\textsuperscript{\rm 6}
    Julie D. Jastrow,\textsuperscript{\rm 6}
    Thomas E. Juenger,\textsuperscript{\rm 3}
    Felix B. Fritschi\textsuperscript{\rm 4}
}
\affiliations{
    \textsuperscript{\rm 1} Department of Electrical and Computer Engineering, University of Florida, Gainesville, FL, USA\\
    \textsuperscript{\rm 2} InTerACT Research Unit, UniLaSalle Bauvais, France \\
    \textsuperscript{\rm 3} Department of Integrative Biology, University of Texas at Austin, Austin, TX, USA\\
    \textsuperscript{\rm 4} Division of Plant Science and Technology, University of Missouri, Columbia, MO, USA\\
    \textsuperscript{\rm 5} College of Natural Sciences, Forestry, and Agriculture, University of Maine, Orono, ME, USA \\
    \textsuperscript{\rm 6} Argonne National Laboratory, Lemont, IL, USA\\
    \{weihuang.xu, guohaoyu, cuiyiming\}@ufl.edu,
    romain.gloaguen@unilasalle.fr,
    azare@ece.ufl.edu,
    jebonnette@utexas.edu,
    jer.cabrera@gmail.com,
    ashish.rajurkar@missouri.edu,
    diane.rowland@maine.edu,
    \{matamala, jdjastrow\}@anl.gov,
    tjuenger@austin.utexas.edu,
    fritschif@missouri.edu

}

\usepackage{bibentry}

\begin{document}

\maketitle

\begin{abstract}
Understanding a plant's root system architecture (RSA) is crucial for a variety of plant science problem domains including sustainability and climate adaptation. Minirhizotron (MR) technology is a widely-used approach for phenotyping RSA non-destructively by capturing root imagery over time. Precisely segmenting roots from the soil in MR imagery is a critical step in studying RSA features. In this paper, we introduce a large-scale dataset of plant root images captured by MR technology. In total, there are over 72K RGB root images across six different species including cotton, papaya, peanut, sesame, sunflower, and switchgrass in the dataset. The images span a variety of conditions including varied root age, root structures, soil types, and depths under the soil surface. All of the images have been annotated with weak image-level labels indicating whether each image contains roots or not. The image-level labels can be used to support weakly supervised learning in plant root segmentation tasks. In addition, 63K images have been manually annotated to generate pixel-level binary masks indicating whether each pixel corresponds to root or not. These pixel-level binary masks can be used as ground truth for supervised learning in semantic segmentation tasks. By introducing this dataset, we aim to facilitate the automatic segmentation of roots and the research of RSA with deep learning and other image analysis algorithms.
\end{abstract}

\section{Introduction}
Plant root systems play important roles in supporting our natural ecosystems, adapting to changes in climate, and ensuring the sustainable production of plants \cite{aidoo2016tolerance,alexander2015novel,ettinger2017competition}.
Phenotyping RSAs is an important component of understanding plant root systems to support and enhance these roles along with advancing many aspects of plant science research \cite{pieruschka2019plant, trachsel2011shovelomics, wasson2012traits}.

\begin{figure}[t]
\centering
\includegraphics[width=0.7\linewidth]{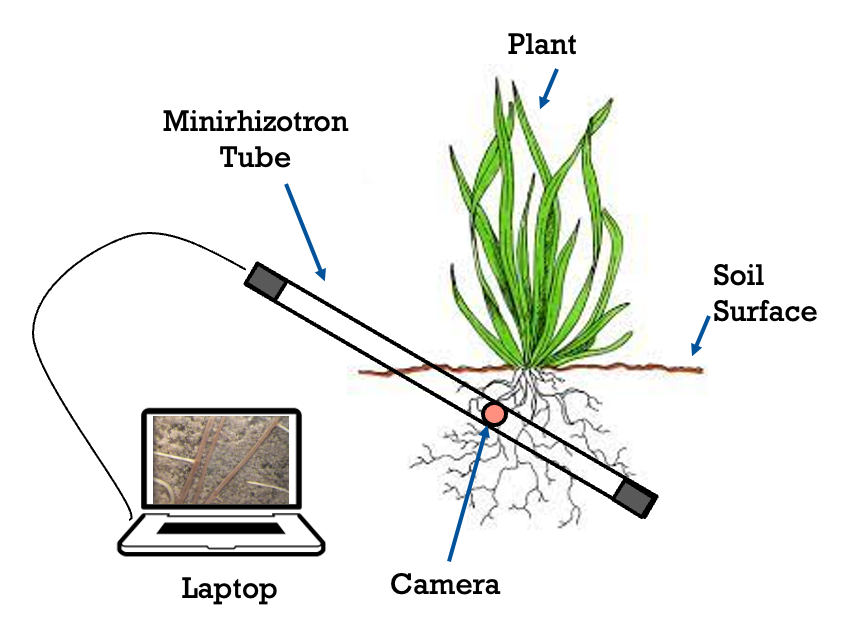}
\caption{Illustration of a MR system. Before planting, transparent tubes are inserted into the soil with the desired angle. A high-resolution camera slides into the MR tubes to take images of the roots that happen to grow along the surface of the tube at different depths over time.}
\label{fig:MR_system} 
\end{figure}

Relative to above-ground plant phenotyping, below-ground field-based plant root phenotyping is challenging. Many common approaches for field-based root phenotyping, such as soil coring and ``shovelomics" \cite{van1979loss, gregory2008plant, trachsel2011shovelomics}, are destructive, laborious, time-consuming, and cannot be carried out in real-time or over-time. 
Minirhizotron (MR) technology \cite{majdi1996root, johnson2001advancing} is one of the few most widely-used approaches for phenotyping RSA non-destructively over time. Generally, before planting, MR transparent tubes are installed in the field at an angle (e.g. $45^{\circ}$) to the soil surface in locations that should eventually be directly under or near to plants of interest. Then, as the plant's root systems grow, a high-resolution camera can be inserted along the tube to capture root images at a variety of depths as shown in Figure \ref{fig:MR_system}. Since the MR tubes remain in the soil during the entire growing period, the camera is able to capture time-series RGB root images providing insight into RSA development. In addition to the development of the whole root structure, MR imagery can be used to observe changes of roots themselves throughout their life cycle such as color, diameter, angle, and length changes.

Precisely segmenting roots from the soil background in MR imagery is a critical step in studying root systems using MR technology. Traditionally, MR images are manually annotated in which users trace along each individual root to indicate the location and diameter. As can be imagined, manually tracing a large number of thin, hard-to-see roots is incredibly tedious and time-consuming. This annotation step is the primary bottleneck limiting the effectiveness of MR technology for large-scale RSA studies. Given the complexity of plant RSA, biological variation, and environmental- and management- impacts on RSA, a huge amount of data is needed for comprehensive RSA studies to lead to statistically reasonable conclusions. Thus, algorithms for high-throughput automatic root segmentation tasks are essential.

Deep learning, and convolutional neural networks specifically, have shown initial great success in plant root segmentation tasks \cite{yasrab2020rootnet,smith2020segmentation}. The features learned by well-trained deep neural networks are more representative and effective for segmenting roots as compared to earlier image processing-based approaches \cite{das2015digital, galkovskyi2012gia, pierret2013ij_rhizo, haralick1987image, lobregt1980three}.
However, one limitation in applying these approaches is the need for large MR image datasets for training that include the soil characteristics and plant species of interest. Transfer learning has been investigated to help alleviate the need for collecting and annotating enormous MR image training sets for each soil type and plant species \cite{xu2020overcoming}. However, much of these investigations have been limited by the lack of public sizeable public MR image datasets due to the fact that collecting and annotating images are extremely tedious, time-consuming and often requires some knowledge and expertise. 

In order to help fill this gap and introduce the computer vision community to this problem domain, we compiled a dataset, named PRMI (Plant Root Minirhizotron Imagery), containing over 72 thousand MR RGB root images across six different species, namely, cotton, papaya, peanut, sesame, sunflower, and switchgrass. These images span a variety of conditions including varying root age, root spatial structures, soil types, and depths under the soil surface. We paired binary image-level labels for each image showing whether the image contains roots along with meta-data such as crop species, collection location, MR tube number, collection time, collection depth, and image resolution. In addition, over 63 thousand of the images have been manually annotated to generate pixel-level binary masks indicating whether the pixels correspond to root or not. Annotating MR root images is challenging, so learning from weak labels is essential, particularly when transitioning a trained method from one location (i.e., soil type and properties) and one plant species (i.e., root characteristics) to another. The primary use of the PRMI dataset is for supervised/weakly-supervised semantic segmentation as well as transfer learning. The RSA features (length, diameter, etc.) can be extracted for scientific purposes after segmentation. We hope this dataset can make contributions to facilitate the automatic segmentation of roots and the research of RSA with deep learning. The dataset is publicly available: \url{https://gatorsense.github.io/PRMI/}.


\section{Related Datasets and Annotation Tools}
There are a few public MR image datasets available as summarized in Table \ref{tab:pub_MRdata}. However, these datasets are relatively small (e.g., the largest collection for one species contains only 400 images) and only three of these datasets provide pixel-level ground truth masks. These datasets have been used to develop image processing-based approaches as well as deep-learning methods for root segmentation tasks. For example, early work on the Peach1 dataset \cite{zeng2006detecting} proposed an image processing-based approach that first extracted a collection of features using linearly-shaped spatial filters. The soybean \cite{wang2019segroot} and the chicory \cite{smith2020segmentation} datasets have been used for developing deep learning segmentation approaches. 

There are several annotation tools that have been used for manually tracing roots. Zeng \textit{et. al.}, who provided the Peach1 dataset \cite{zeng2006detecting}, developed Rootfly for root tracing. Le Bot \textit{et. al.} developed a software DART \cite{le2010dart} to analyze RSA. Commercial tools are also available such as RootSnap (CID Bioscience) and WinRHIZO Tron (Regent Instrument) \cite{bauhus1999evaluation}. Furthermore, general image processing software such as Photoshop and ImageJ have also been used for root tracing \cite{abramoff2004image}. Automated tools were also developed such as EZ-RHIZO \cite{armengaud2009ez}, RootNav \cite{yasrab2019rootnav}, and RootGraph \cite{cai2015rootgraph}.

\begin{table}[t]
\centering
  \begin{tabular}{cccc}
    \toprule
    Dataset  & Num of images(Train/Test) & Label \\
    \midrule
    Peach1$^*$ & 50/200  & Pixel label\\
    Peach2$^*$ & 200/200 & Image Label\\
    maple1$^*$ &120/120 & Image Label\\
    magnolia1$^*$ & 100/100 & Image Label\\
    soybean$^\S$ & 39/13 & Pixel label \\
    chicory$^\ddagger$ & 38/10 & Pixel label\\
    \bottomrule
  \end{tabular}
  \begin{tablenotes}
    \begin{small}
        \item $^*$ \cite{zeng2006detecting}; $^\S$ \cite{wang2019segroot}; $^\ddagger$ \cite{smith2020segmentation}
    \end{small}
  \end{tablenotes}
  \caption{Other Public MR Datasets}
  \label{tab:pub_MRdata}
\end{table}

\begin{figure*}[t]
\begin{center}
    \subfloat[Cotton]{%
      \includegraphics[width=0.16\linewidth]{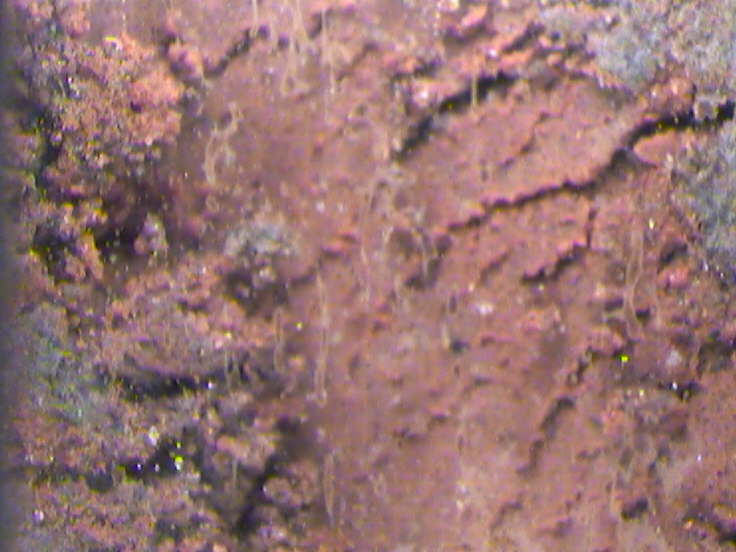}} \hfill
    \subfloat[Papaya]{%
      \includegraphics[width=0.16\linewidth]{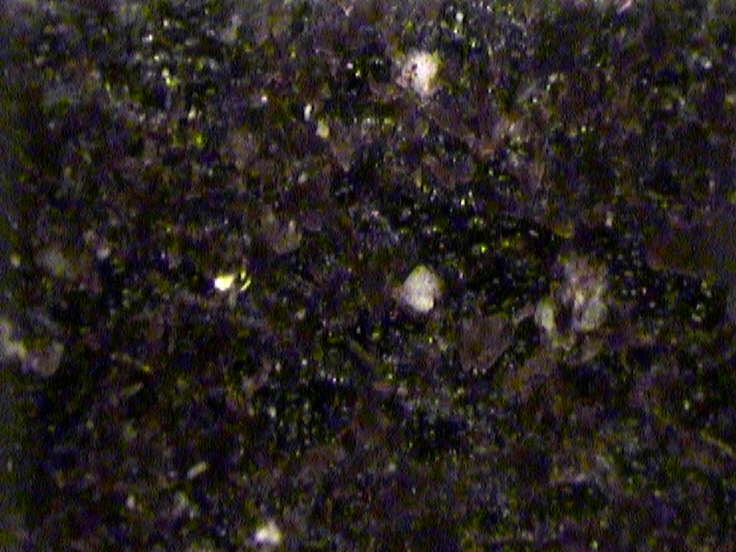}} \hfill
    \subfloat[Peanut]{%
      \includegraphics[width=0.16\linewidth]{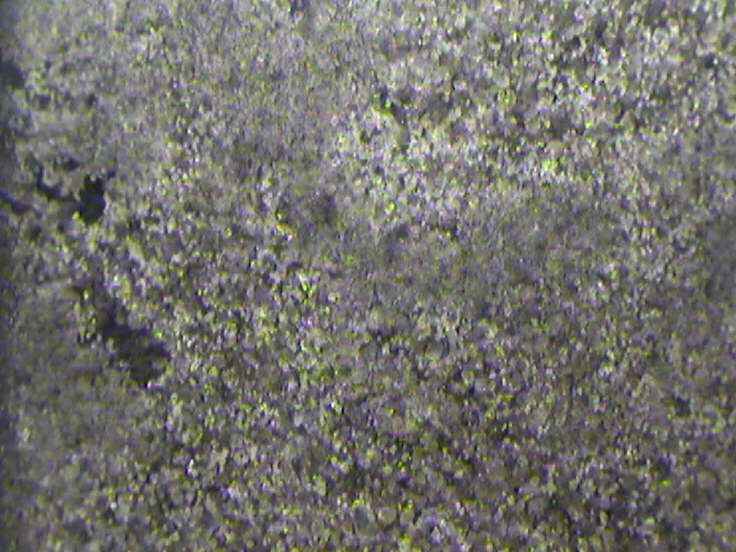}} \hfill
    \subfloat[Sesame]{%
      \includegraphics[width=0.16\linewidth]{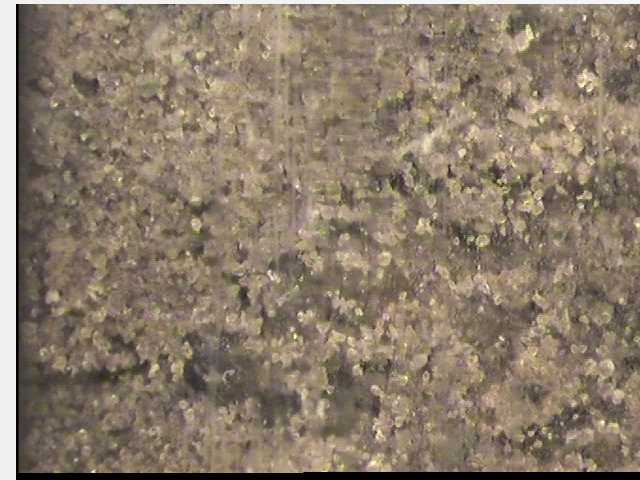}} \hfill
    \subfloat[Sunflower]{%
      \includegraphics[width=0.16\linewidth]{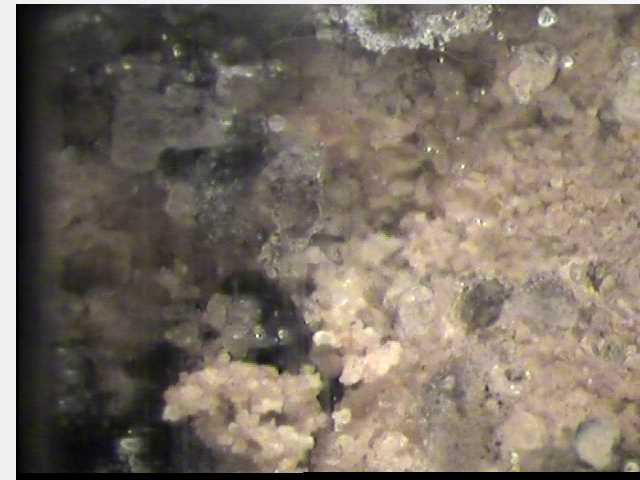}} \hfill
    \subfloat[Switchgrass]{%
      \includegraphics[width=0.12\linewidth, angle=90]{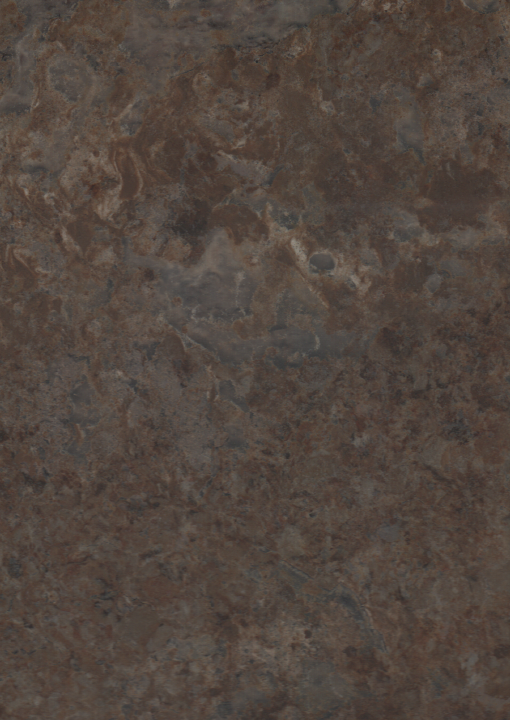}} \hfill
      \\[-2ex]
      
    \subfloat{%
      \includegraphics[width=0.16\linewidth]{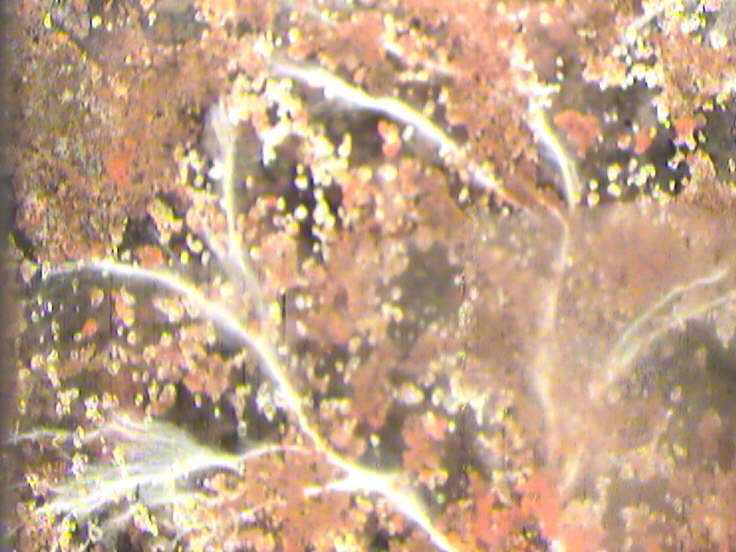}} \hfill
    \subfloat{%
      \includegraphics[width=0.16\linewidth]{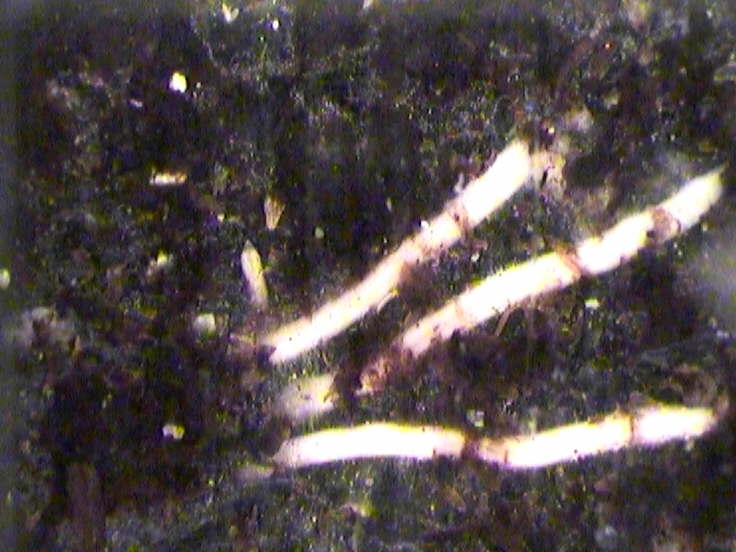}} \hfill
    \subfloat{%
      \includegraphics[width=0.16\linewidth]{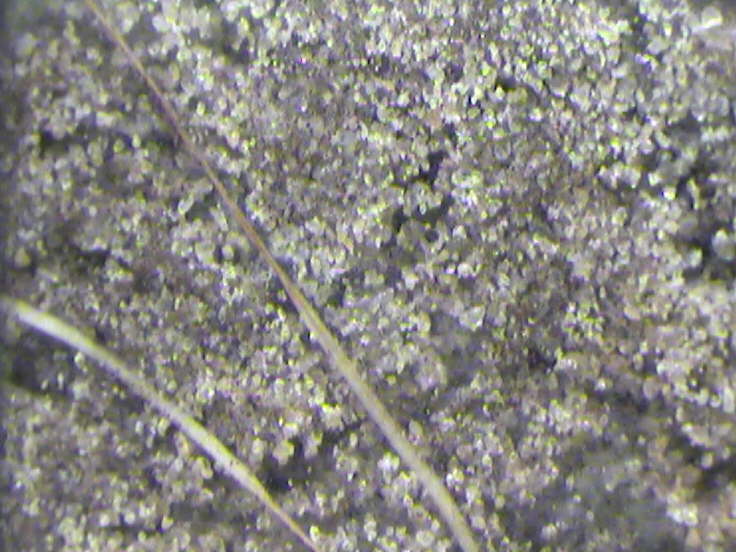}} \hfill
    \subfloat{%
      \includegraphics[width=0.16\linewidth]{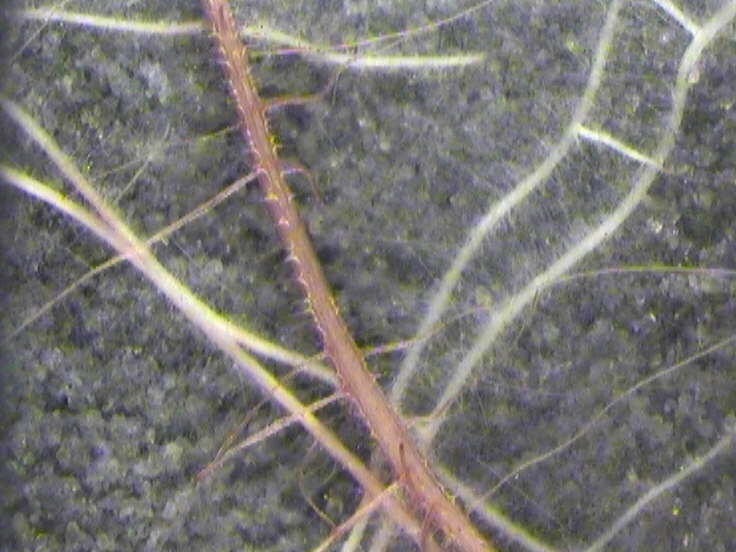}} \hfill
    \subfloat{%
      \includegraphics[width=0.16\linewidth]{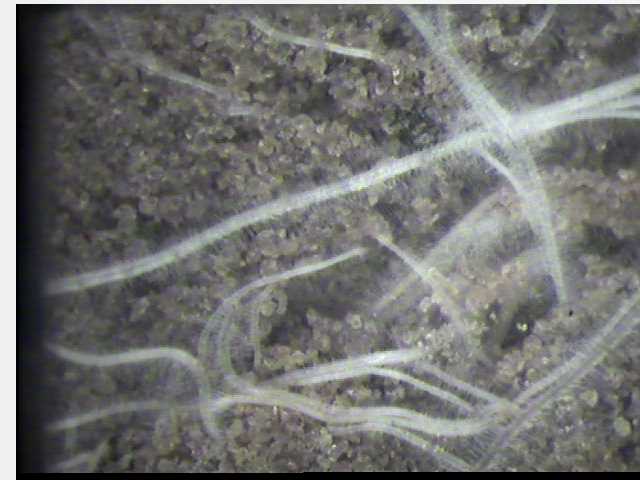}} \hfill
    \subfloat{%
      \includegraphics[width=0.12\linewidth, angle=90]{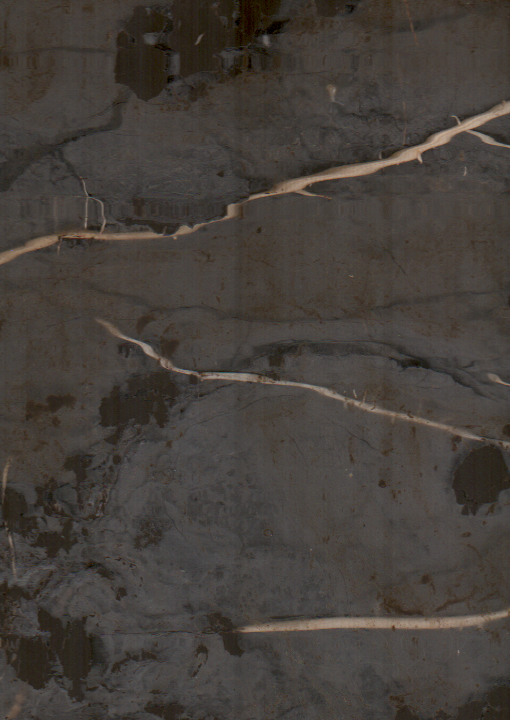}} \hfill
      \\[-2ex]
      
    \subfloat{%
      \fbox{\includegraphics[width=0.148\linewidth]{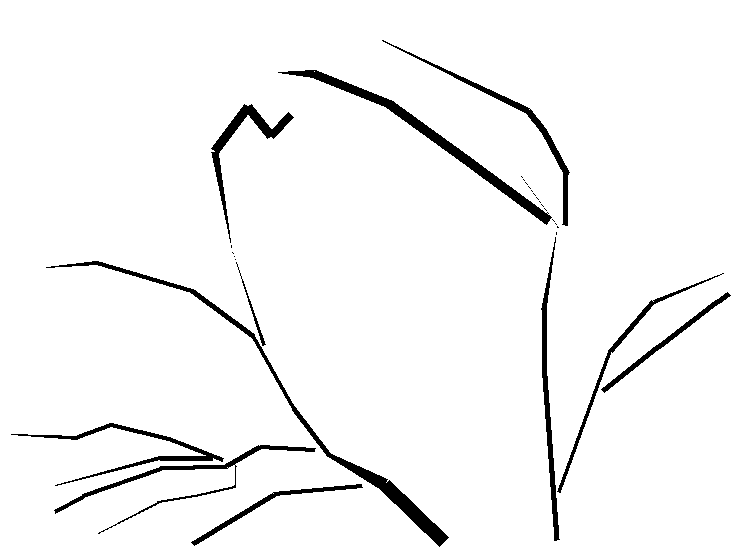}}} \hfill
    \subfloat{%
      \fbox{\includegraphics[width=0.148\linewidth]{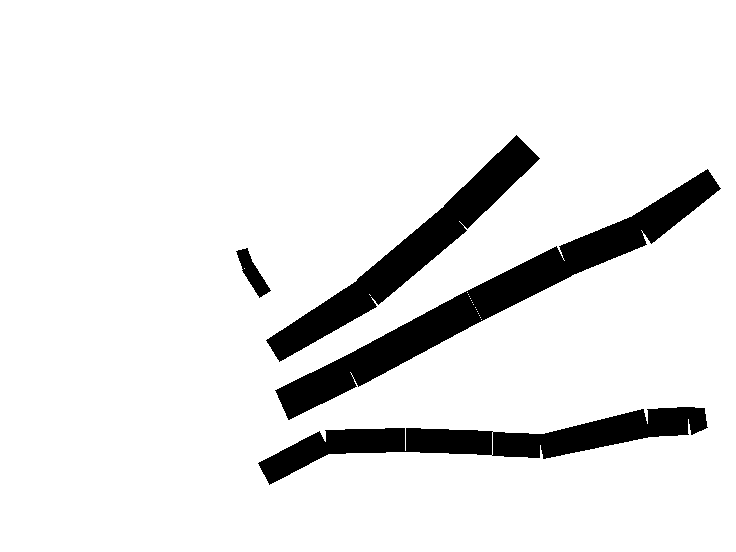}}} \hfill
    \subfloat{%
      \fbox{\includegraphics[width=0.148\linewidth]{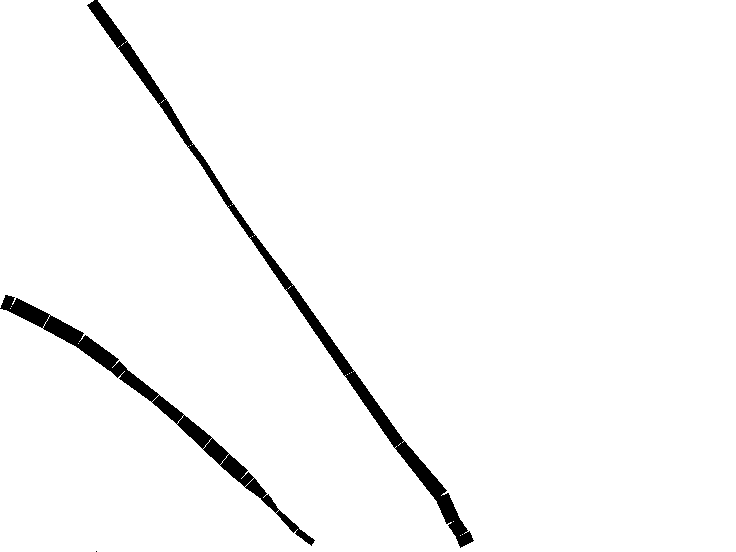}}} \hfill
    \subfloat{%
      \fbox{\includegraphics[width=0.148\linewidth]{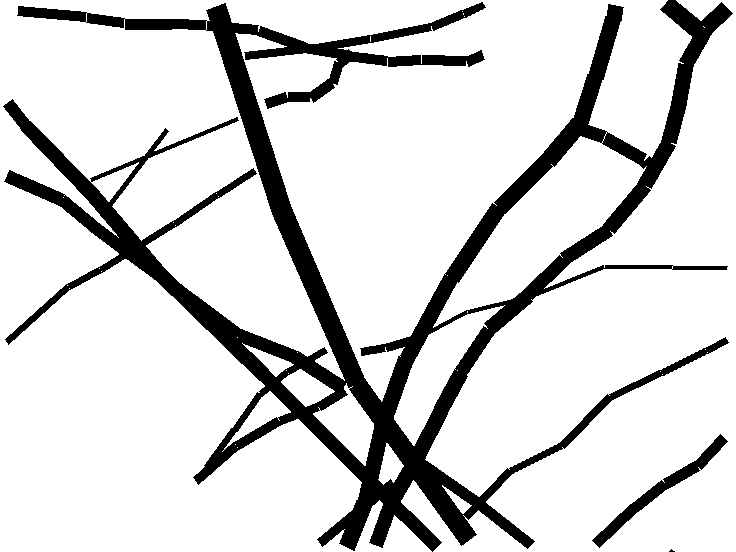}}} \hfill
    \subfloat{%
      \fbox{\includegraphics[width=0.148\linewidth]{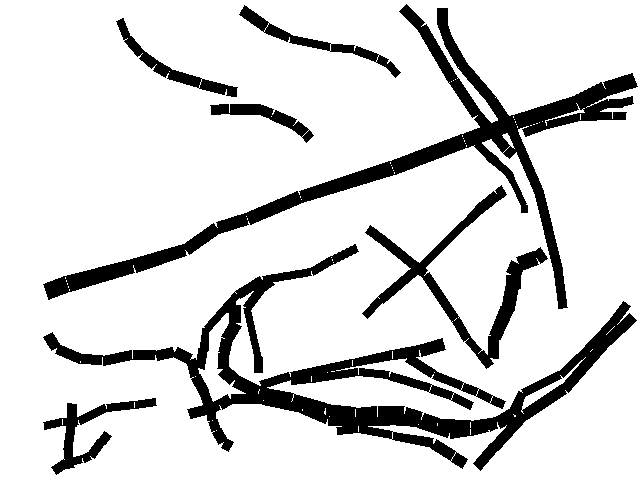}}} \hfill
    \subfloat{%
      \fbox{\includegraphics[width=0.11\linewidth, angle=90]{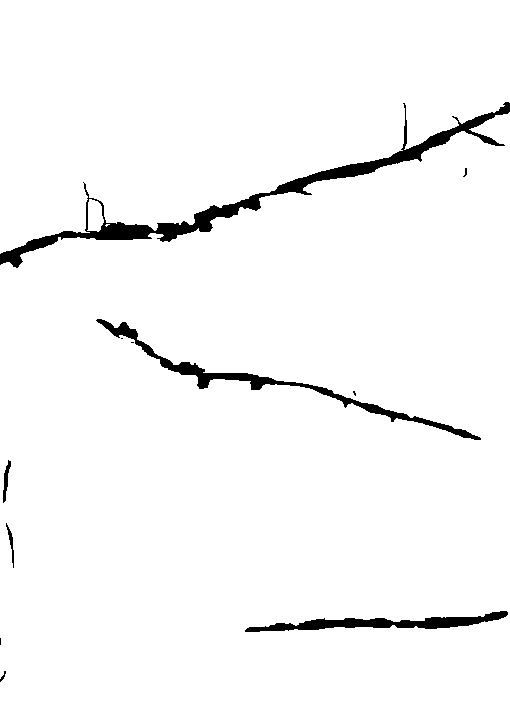}}} \hfill
\end{center}
  \caption{Examples of selected raw root MR images for each species and manually annotated ground truth masks. (a) Cotton, (b) Papaya, (c) Peanut, (d) Sesame, (e) Sunflower, and (f) Switchgrass (rotated by $90^{\circ}$). 1st Rows shows the images have no roots (only soil background). 2nd row shows the images have roots. 3rd rows shows the paired binary masks.}
  \label{fig:selected_imgs} 
\end{figure*}

\section{Image Collection}
\label{image_collect}
In our PRMI dataset, MR technology was used to collect root images across cotton, papaya, peanut, sesame, sunflower, and switchgrass. For each species, root images were captured from multiple plants and multiple MR tubes across a variety of depths over different time periods and varying soil and environmental conditions. Many of the root images are related to each others in time or depth. For example, for a specific species, images were captured from the same MR tube and the same depth across times (i.e., a time series of MR imagery of particular plant root architectures). Thus, some of these images will contain the same root portions at different ages, sizes, shapes and colors. 
Although all the root images of different species were collected by MR technology, the specifics of the MR technology were different for each sub-collection. For example, the camera used for each species was distinct, resulting in different image resolutions, dots per inch (DPI), and color profiles. For each species, there were images containing plant roots as well as images containing purely soil and background. Some example images for each species are shown in the Figure \ref{fig:selected_imgs}.

The details of the sub-collections of each species are shown in Table {\ref{tab:data_stat}}, including the dpi, resolution, number of locations, number of collection tubes, number of root images, number of non-root images, the number of images in train/val/test sets, and the total number of images. To maintain the original appearance of the root images, we compiled the dataset using raw root images in their original format without any pre-processing such as re-sizing or any normalization.  

\begin{table*}[h]
\begin{threeparttable}
\centering
\begin{tabular}{c|c|c|c|c|c|c|c|c|c|c}
\toprule
    \multirow{2}{*}{Species} & \multirow{2}{*}{DPI} & \multirow{2}{*}{Size} & Num of  & Num of  & Num of & Num of non-& \multirow{2}{*}{Train} & \multirow{2}{*}{Val} & \multirow{2}{*}{Test} & \multirow{2}{*}{Total} \\
    &  & & locations & tubes & root images & root images  & & & & \\
\midrule
    Cotton & 150 & $736\times552$ & 1 & 12 & 918 & 1494 & 1271 & 564 & 577 & 2412 \\
    
    Papaya & 150 & $736\times552$ & 2 & 6 & 487 & 59 & 282 & 131 & 133 & 546 \\
    
    Peanut & 120 & $640\times480$ & 1 & 32 & 8508 & 8534 & 10087 & 3413 & 3542 & 17042 \\
    
    Peanut & 150 & $736\times552$ & 2 & 24 & 11147 & 8478 & 11485 & 3347 & 4793 &19625 \\
    
    Sesame & 120 & $640\times480$ & 4 & 11 & 1460 & 700 & 1438 & 318 & 404 & 2160 \\
    
    Sesame & 150 & $736\times552$ & 2 & 24 & 7923 & 6423 & 8637 & 2625 & 3084 & 14346 \\
    
    Sunflower & 120 & $640\times480$ & 1 & 16 & 1646 & 2254 & 2211 & 722 & 967 & 3900 \\
    
    Switchgrass & 300 & $510\times720$ & 3 & 72 & 3465 & 9072 & 11272$^*$ & 665 & 600 & 12537 \\
\bottomrule
\end{tabular}
\begin{tablenotes}
  \small
  \item $^*$ 2647 images in the training set have pixel-level annotation and the remaining 8625 images only have image-level annotation.
\end{tablenotes}
\caption{The summary of MR root images for each species in PRMI dataset.}
\label{tab:data_stat}
\end{threeparttable}
\end{table*}

\section{Image Annotation}
Each MR image in this collection is paired with image-level annotations and more than 63 thousand of the images are also paired with pixel-level annotation. Next, we will introduce the annotation details.

\subsection{Image-level Annotation}
To enable a more comprehensive understanding of the dataset, meta-data for each MR image is recorded and provided in a JSON file (details can be found in the appendix B). This meta-data includes crop species, collection location, MR tube number, collection time, collection depth, and sensor DPI. More importantly, we manually annotate each MR image with an binary image-level label indicating whether the image contains roots. This image-level annotation can be used in a weakly-supervised learning framework for plant root segmentation tasks, which is one of the major proposed use of this data. 

\subsection{Pixel-level Annotation}
Each pixel-level annotation is a segmentation mask containing binary values indicating the class (root/soil) for each pixel. Selected pixel-level segmentation masks for each species are shown in Figure \ref{fig:selected_imgs}. All the pixel-level masks were generated by technicians manually labeling pixels in the images. However, different annotation methods were used to generate masks for switchgrass and all the other species. 

\textbf{Pixel-level annotation for all species other than switchgrass}.
WinRHIZO Tron software was used to manually trace the roots in images of cotton, papaya, peanut, sesame, and sunflower. The pixel-level segmentation masks were generated by drawing rectangular boxes to highlight the area of root pixels in the software. To make the annotation as accurate as possible, multiple rectangular boxes of different sizes were used for each single piece of root to capture any variance in diameter. The coordinates of four vertexes of all rectangular boxes were recorded for each root image. Then, the binary masks were generated by reconstructing and aligning these boxes in a blank image of the same size as the corresponding RGB root images. 

\textbf{Pixel-level annotation for the switchgrass}. 
There are 3912 out of 12537 switchgrass images paired with pixel-level segmentation masks. Given the fine, narrow switchgrass roots and the size of these images, it is extremely time-consuming to manually annotate each pixel. To improve the efficiency of the annotation process, two different annotation methods were used: a) technicians annotated images on superpixels generated by running the Simple Linear Iterative Clustering (SLIC) algorithm \cite{achanta2010slic} on raw images; and b) technicians refined the predicted segmentation masks generated by passing raw images to pre-trained U-net. Details can be found in appendix B.

\section{Dataset Split} \label{sec:data_split}
For each species, the root images were collected from different locations, dates, tubes, and depths. To ensure that the training, validation, and testing data share a similar data distribution while ensuring that testing data does not contain the same tubes and roots used for training, we split the data into train, validation and test sets based on the tubes. 

We randomly selected 60\% of tubes for training, 20\% for validation, and the remaining 20\% for test for each species other than switchgrass. In terms of switchgrass root images, to ensure confidence in the accuracy of segmentation masks in the test set, all the images with manually generated ground-truth were placed in the test set. The training set was generated by randomly selected 80\% of tubes with AI-guided annotation with the remaining compiled in the validation set. In addition, all the images with only image-level annotations were placed in the training set to support weakly-supervised segmentation. The total number of images for train, validation and test sets are shown in Table \ref{tab:data_stat}.

The dataset could also be split according to other factors such as dates, depths, and locations for specific research questions. For example, researchers can split the dataset based on dates to study the changes of roots over time. We paired image-level labels with meta-data for each image, which can be easily used to sort and split dataset based on different attributes.

\section{Experiments and Results}
The primary proposed use of this PRMI dataset is for root segmentation tasks. To generate baseline benchmark results and show the challenges of this dataset, we trained a U-net \cite{ronneberger2015u} for supervised learning and IRNet \cite{ahn2019weakly} for weakly-supervised learning on each sub-collection species to show the segmentation results. All the models were trained on training set for 300 epochs using a single Nvidia A100 GPU. For each sub-collection, the model performing best on the validation set will be used to evaluate the segmentation performance on the test set. The Intersection-Over-Union (IoU) and F1 score were calculated as evaluation metrics. More details about training can be found in the appendix D.

Table {\ref{tab:evaluation_E150}} shows the F1 and IoU scores for each sub-collection. In general, the dataset is quite challenging as the supervised model got around 35\% average IoU for all species. This dataset is even more challenging for CAM-based weakly-supervised model since the CAM methods focus on the most different parts in the images which could be complicated soil conditions underground instead of roots when roots are rare and sparse. In addition, the number of root pixels and the number of soil pixels are highly imbalanced resulting in the fact that model could be easily biased towards soil features instead of root features. Some species, such as cotton and sunflower, have relative less images having roots compared with other species. With these facts, we believe this dataset can be helpful for researchers to build and develop more robust and advanced algorithms for automatic root segmentation tasks.

\begin{table}[bt]
\begin{threeparttable}
\centering
\begin{tabular}{c|c|c|c|c}
\toprule
\multirow{2}{*}{Species-DPI} & \multicolumn{2}{c|}{U-net} & \multicolumn{2}{c}{IRNet}  \\
\cline{2-5}
\rule{0pt}{10pt} & IoU & F1 & IoU & F1 \\
\midrule
Cotton-150 & 4.8\% & 0.092 & 0.8\% & 0.016 \\
Papaya-150 & 56.0\% & 0.718 & 14.4\% & 0.252 \\
Peanut-150 & 36.8\% & 0.538 & 2.6\% & 0.055\\
Peanut-120 & 61.9\% & 0.765 & 10.6\% & 0.192\\
Sesame-150 & 20.6\% & 0.341 & 4.4\% & 0.084 \\
Sesame-120 & 25.9\% & 0.411 & 7.2\% & 0.134 \\
Sunflower-120 & 29.7\% & 0.458 & 2.3\% & 0.045 \\
Switchgrass-300 & 40.2\% & 0.574 & 2.2\% & 0.043 \\
\bottomrule
\end{tabular}
\caption{The IoU and F1 scores for each species in PRMI dataset.}
\label{tab:evaluation_E150}
\end{threeparttable}
\end{table}

\section{Conclusion}
In this work, we present PRMI, a large-scale dataset of diverse plant root images captured by MR technology across different species in a variety of conditions. This dataset is very well suited to study supervised/weakly-supervised semantic segmentation, transfer learning, one- or few-shot learning, and data distribution shifts. Through this work, we hope the PRMI dataset can bring more attention of computer vision community to this problem domain and facilitate the research of weak-learning mechanisms and/or transfer-learning mechanisms that can be applied with very limited data and significant changes to the data distribution.

\appendix

\section{Appendix A: Image Collection}
\begin{figure*}[t]
\begin{center}
    \subfloat[Cotton]{%
      \includegraphics[width=0.16\linewidth]{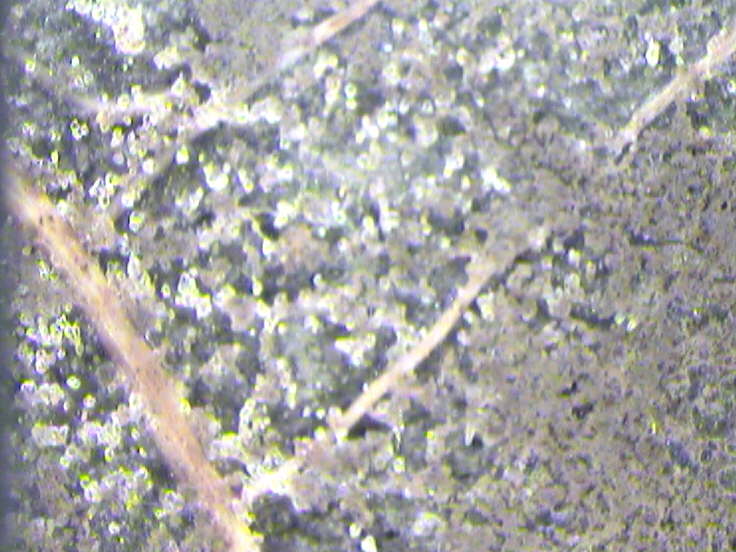}} \hfill
    \subfloat[Papaya]{%
      \includegraphics[width=0.16\linewidth]{./PAPAYASHADEPRIMING1_T092_L011_2013.07.19_134254_002_CRI.jpg}} \hfill
    \subfloat[Peanut]{%
      \includegraphics[width=0.16\linewidth]{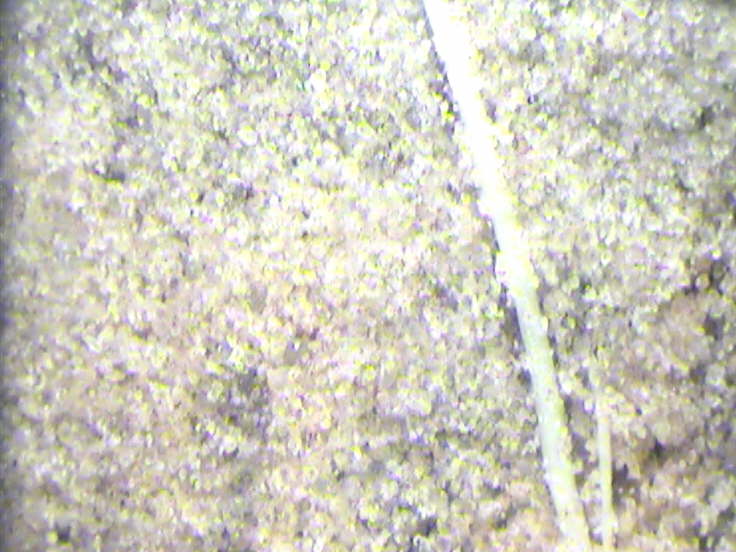}} \hfill
    \subfloat[Sesame]{%
      \includegraphics[width=0.16\linewidth]{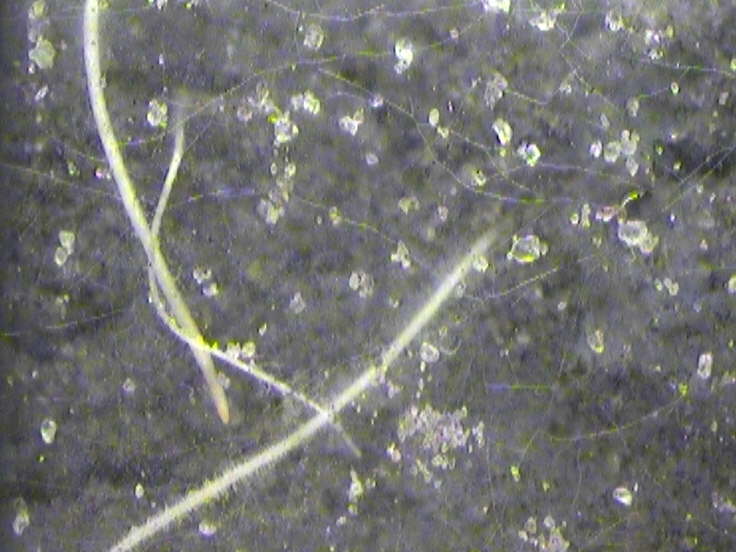}} \hfill
    \subfloat[Sunflower]{%
      \includegraphics[width=0.16\linewidth]{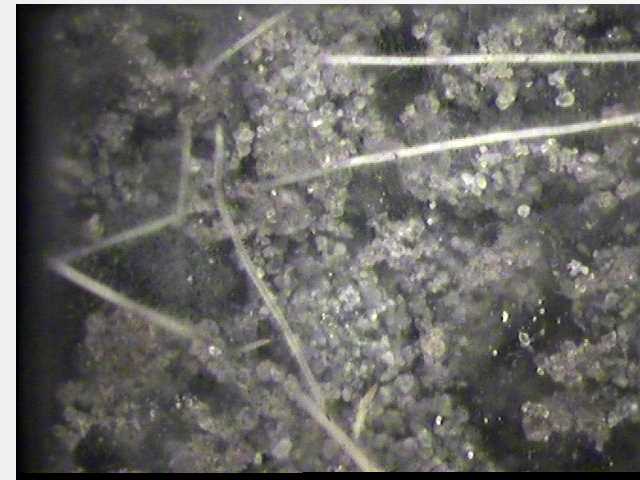}} \hfill
    \subfloat[Switchgrass]{%
      \includegraphics[width=0.12\linewidth, angle=90]{./DOE.S300_T312_L2_09.08.17_114706_1_FRMI_1_5.png}} \hfill
      \\[-2ex]
    \subfloat{%
      \includegraphics[width=0.16\linewidth]{./QUINCYCOTTON2012_T006_L016_2012.08.06_134323_003_AMC.jpg}} \hfill
    \subfloat{%
      \includegraphics[width=0.16\linewidth]{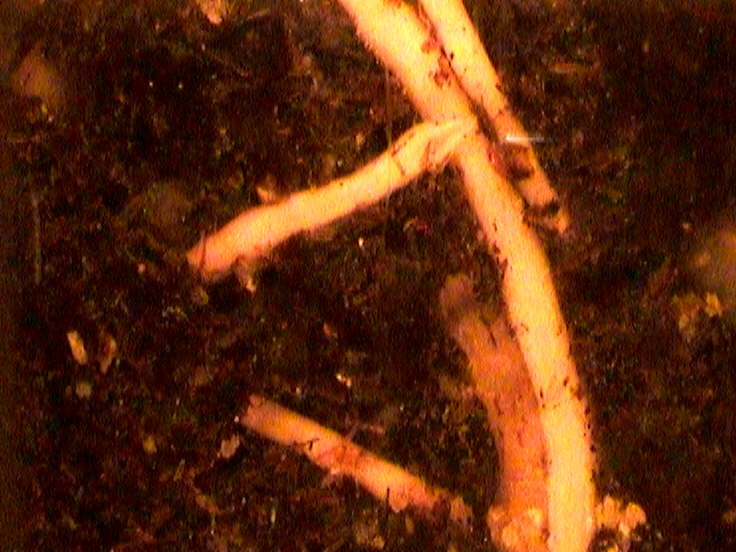}} \hfill
    \subfloat{%
      \includegraphics[width=0.16\linewidth]{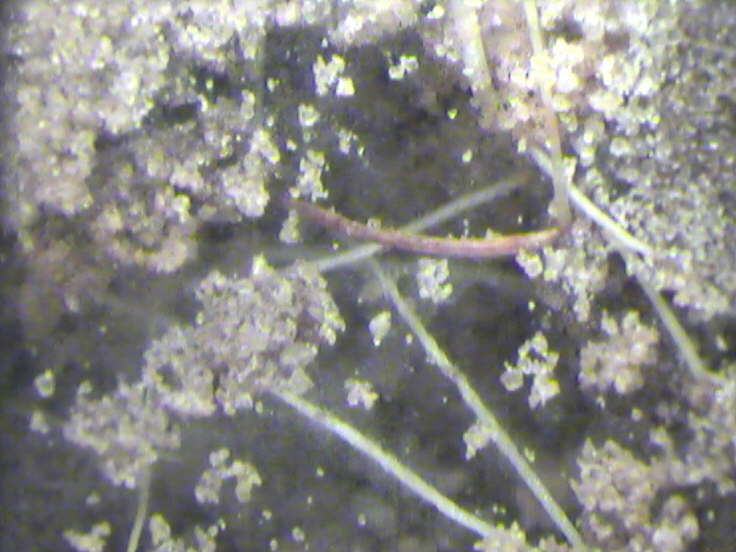}} \hfill
    \subfloat{%
      \includegraphics[width=0.16\linewidth]{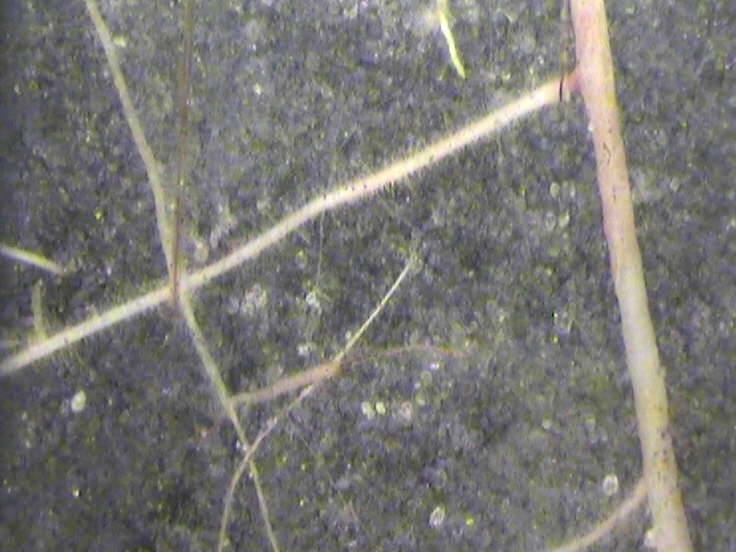}} \hfill
    \subfloat{%
      \includegraphics[width=0.16\linewidth]{./ERINSUNFLOWER2017_T004_L057_2017.07.18_064953_003_EED.jpg}}\hfill
    \subfloat{%
      \includegraphics[width=0.12\linewidth, angle=90]{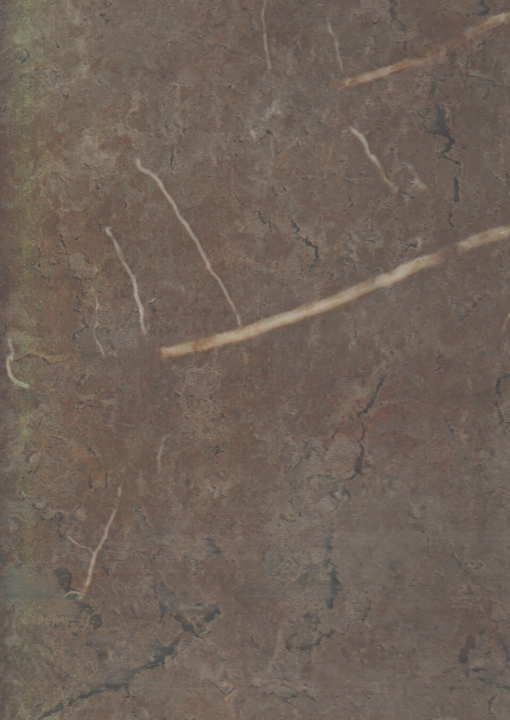}} \hfill
      \\[-2ex]
    \subfloat{%
      \includegraphics[width=0.16\linewidth]{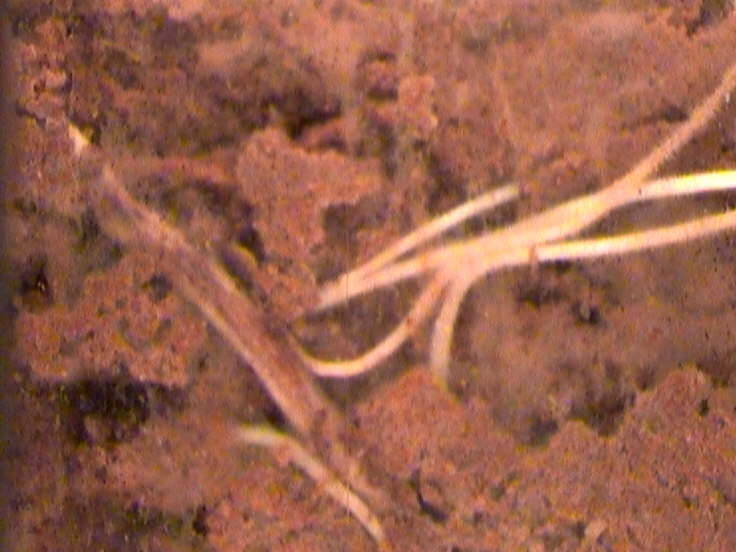}} \hfill
    \subfloat{%
      \includegraphics[width=0.16\linewidth]{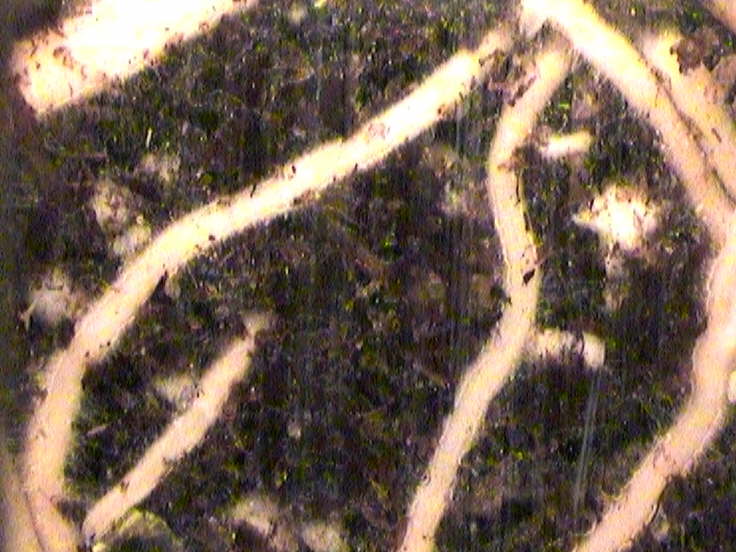}} \hfill
    \subfloat{%
      \includegraphics[width=0.16\linewidth]{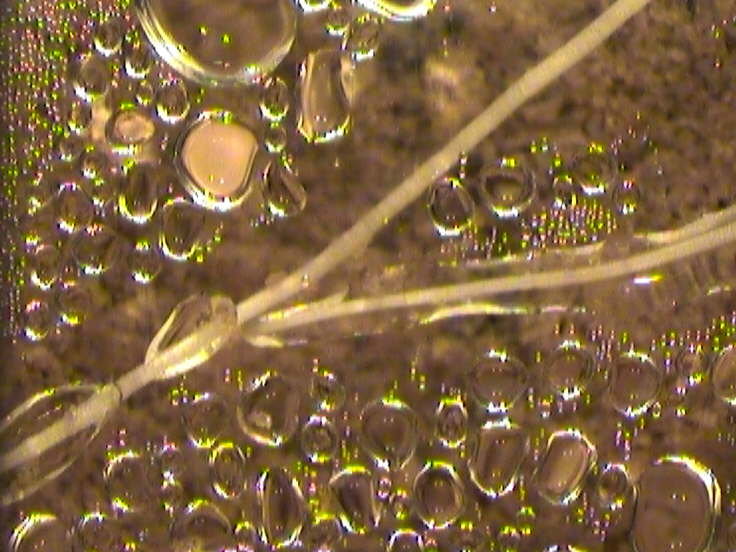}} \hfill
    \subfloat{%
      \includegraphics[width=0.16\linewidth]{./2014SESAMECITRA_T001_L051_2014.08.15_085502_004_AMC.jpg}} \hfill
    \subfloat{%
      \includegraphics[width=0.16\linewidth]{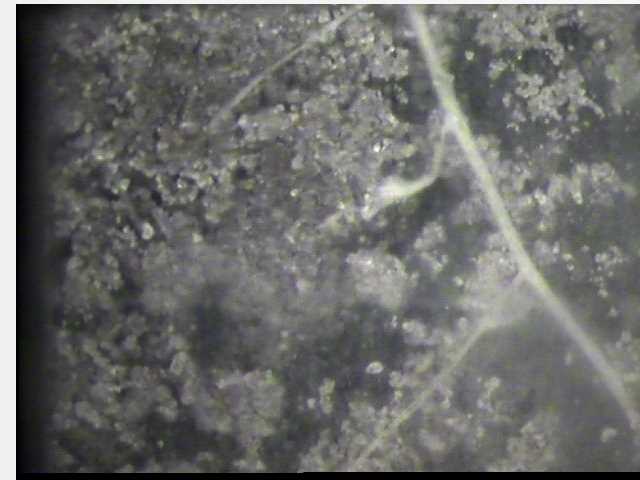}} \hfill
    \subfloat{%
      \includegraphics[width=0.12\linewidth, angle=90]{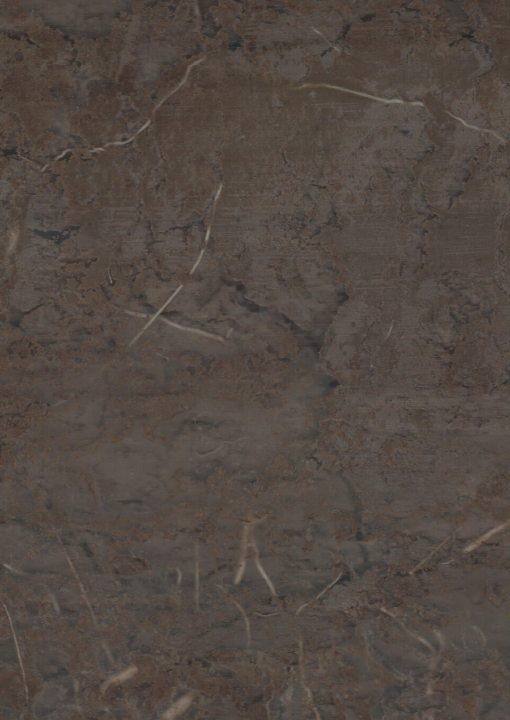}} \hfill
      \\[-2ex]
    \subfloat{%
      \includegraphics[width=0.16\linewidth]{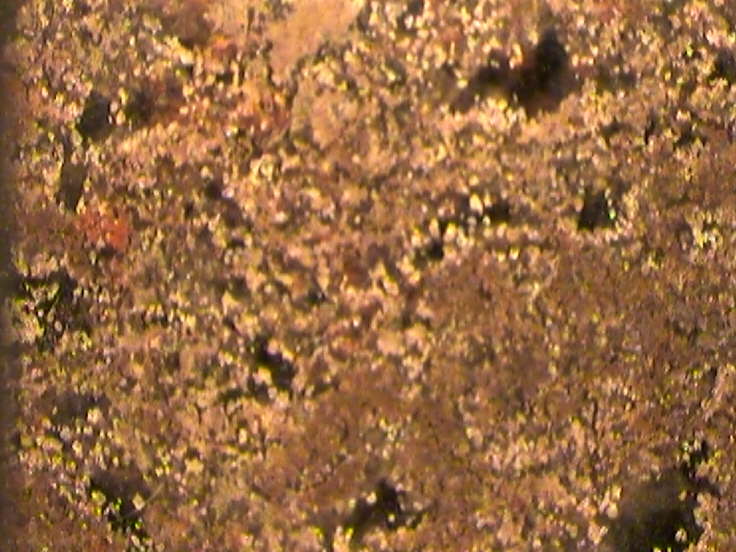}} \hfill
    \subfloat{%
      \includegraphics[width=0.16\linewidth]{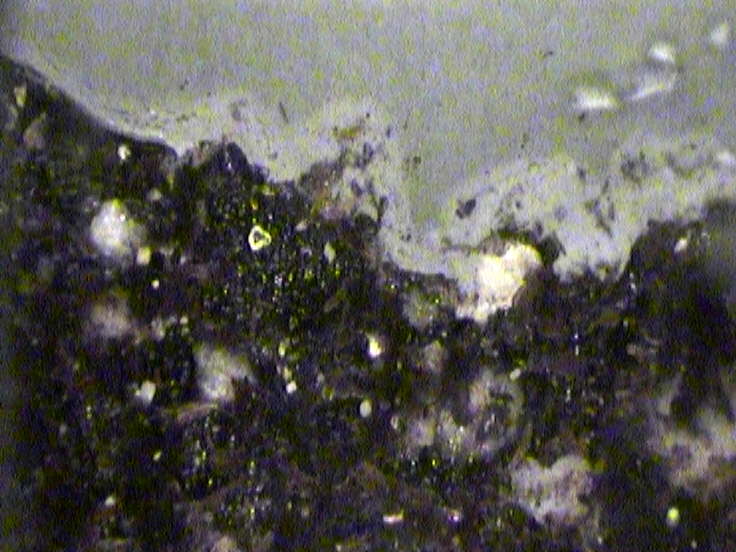}} \hfill
    \subfloat{%
      \includegraphics[width=0.16\linewidth]{./SHELTERPEANUTS2015_T001_L017_2015.06.11_120723_004_BAZ.jpg}} \hfill
    \subfloat{%
      \includegraphics[width=0.16\linewidth]{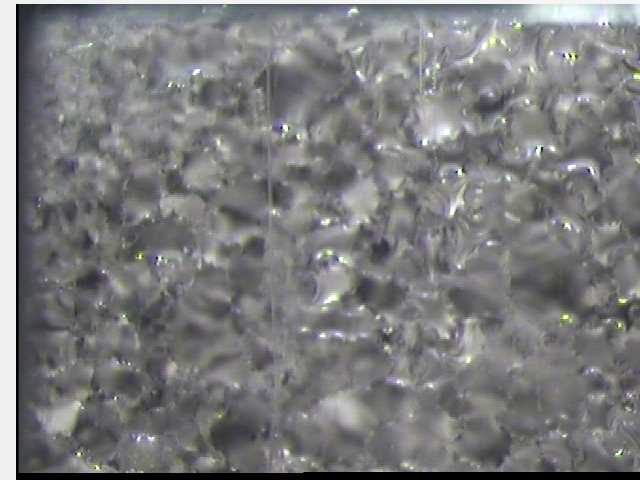}} \hfill
    \subfloat{%
      \includegraphics[width=0.16\linewidth]{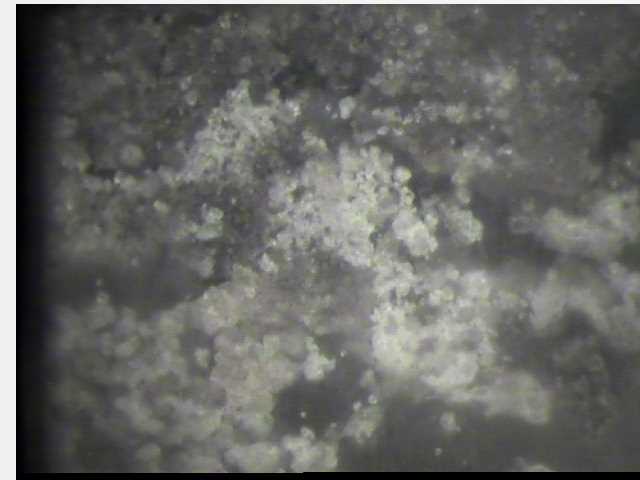}} \hfill
    \subfloat{%
      \includegraphics[width=0.12\linewidth, angle=90]{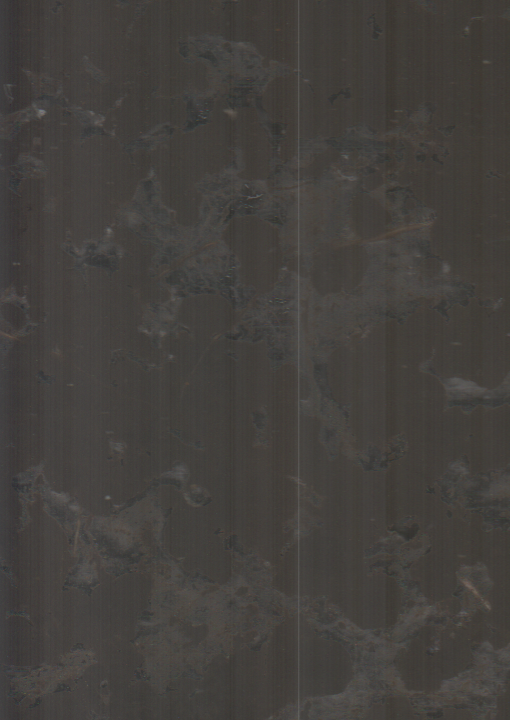}} \hfill
      \\[-2ex]
    \subfloat{%
      \includegraphics[width=0.16\linewidth]{./QUINCYCOTTON2012_T014_L089_2012.08.06_151124_003_AMC.jpg}} \hfill
    \subfloat{%
      \includegraphics[width=0.16\linewidth]{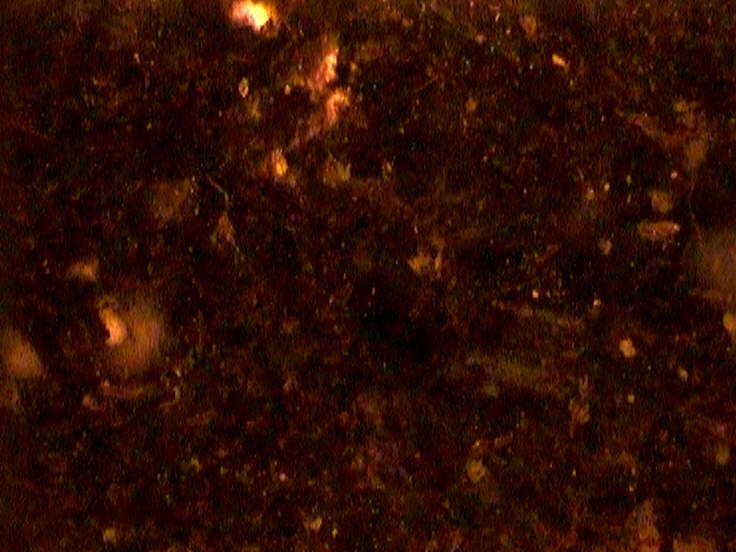}} \hfill
    \subfloat{%
      \includegraphics[width=0.16\linewidth]{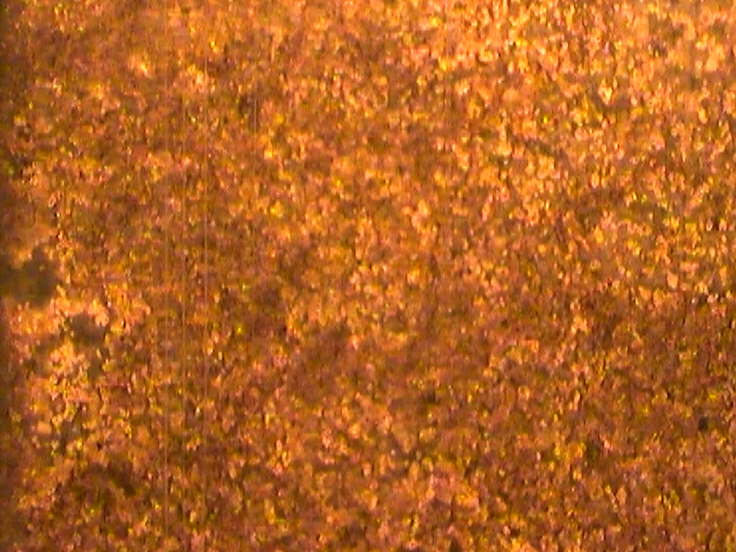}} \hfill
    \subfloat{%
      \includegraphics[width=0.16\linewidth]{./2018SESAME_T011_L013_2018.07.17_102426_002_RS.jpg}} \hfill
    \subfloat{%
      \includegraphics[width=0.16\linewidth]{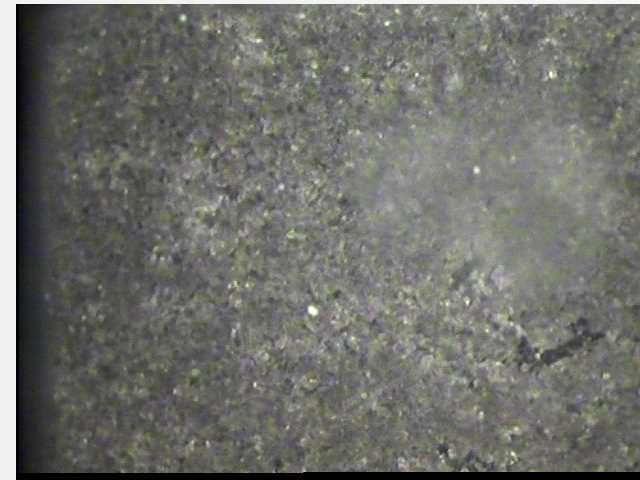}} \hfill
    \subfloat{%
      \includegraphics[width=0.12\linewidth, angle=90]{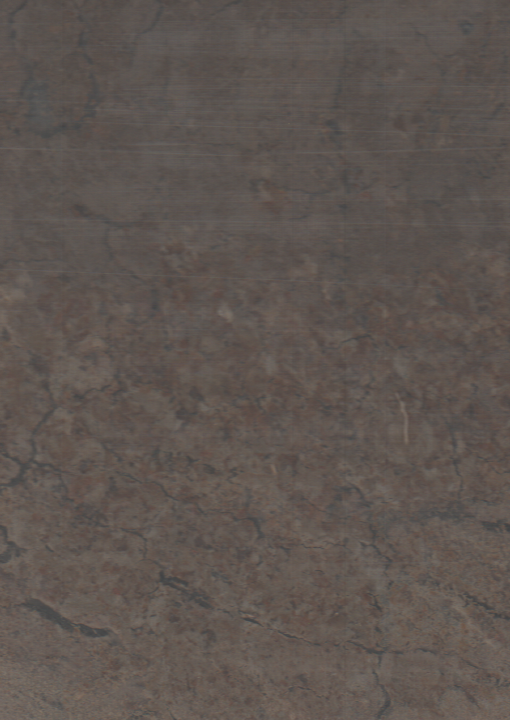}} \hfill
      \\[-2ex]
    \subfloat{%
      \includegraphics[width=0.16\linewidth]{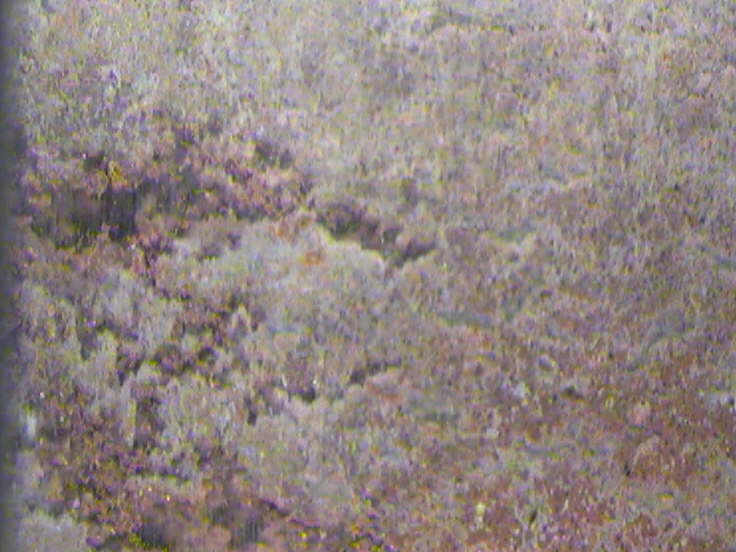}} \hfill
    \subfloat{%
      \includegraphics[width=0.16\linewidth]{./PAPAYASHADEPRIMING1_T101_L009_2013.07.12_142802_001_CIV.jpg}} \hfill
    \subfloat{%
      \includegraphics[width=0.16\linewidth]{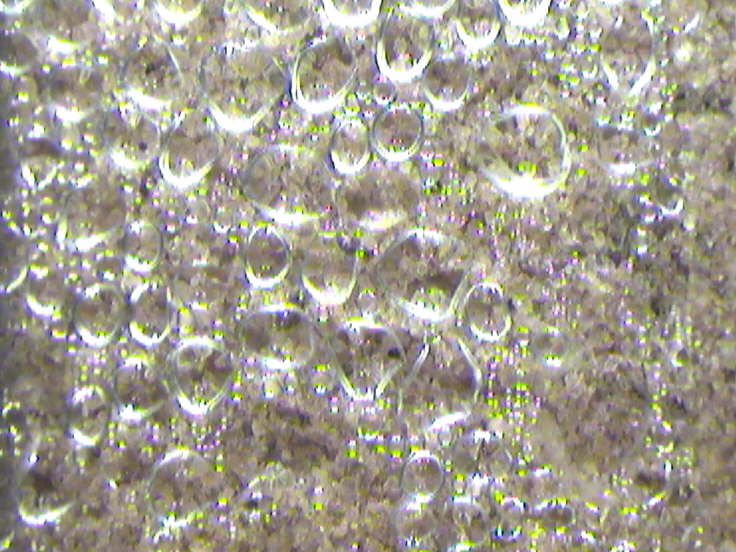}} \hfill
    \subfloat{%
      \includegraphics[width=0.16\linewidth]{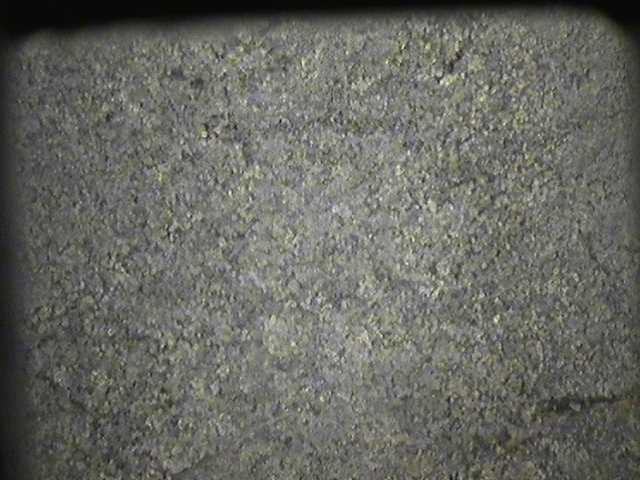}} \hfill
    \subfloat{%
      \includegraphics[width=0.16\linewidth]{./ERINSUNFLOWER2017_T007_L024_2017.07.05_085638_002_EED.jpg}} \hfill
    \subfloat{%
      \includegraphics[width=0.12\linewidth, angle=90]{./DOE300_T131_L5_30.09.18_124654_1_CLMB_1_4.png}} \hfill
\end{center}
  \caption{Examples of selected raw root MR images for each species. (a) Cotton, (b) Papaya, (c) Peanut, (d) Sesame, (e) Sunflower, and (f) Switchgrass (rotated by $90^{\circ}$). Rows 1-3 show the images have roots and rows 4-6 show the images have no roots (only soil background).}
  \label{Selected raw imgs} 
\end{figure*}

Although all the root images of different species were collected by MR technology, the specifics and characteristics of the MR images were different for each sub-collection. For each species, root images were captured from multiple plants and multiple MR tubes across a variety of depths over different time periods and varying soil and environmental conditions. For each species, there were images containing plant roots as well as images containing purely soil and background. To help better understand the varieties of root images as well as different soil background, some example images for each species are shown in the Figure \ref{Selected raw imgs}. The top three rows show the images containing roots, and the bottom three rows show non-root images with different soil types and other background components (e.g., water bubbles and condensation). 

Different from other species, the raw switchgrass images were captured by a high-resolution camera with a resolution of $2550 \times 2160$. Due to the drastically large size compared to the other sub-collections, we divided each image into 15 sub-images of size $510 \times 720$ without overlap. The row index and column index for each sub-image are saved in both image names and the image-level labels such that the original high-resolution images can be reconstructed if needed. After dividing, the switchgrass sub-collection has 3465 sub-images with roots, 9072 sub-images without roots, resulting in a total of 12,537 sub-images in total. 

In addition, all of the images other than switchgrass were saved in JEPG format and named using the following convention: \textit{Species\_tube\_depth\_date\_time\_location\_DPI}. For switchgrass images, all the images were saved in PNG format and named using the following convention: \textit{Species\_tube\_depth\_date\_time\_location\_DPI\_row-index\_column-index}.

\section{Appendix B: Image Annotation}
\subsection{Image-level Annotation}
We provided meta-data for each image including crop species,  collection  location,  MR  tube  number, collection time, collection depth, and sensor DPI.
The image-level annotations are provided in a JavaScript Object Notation (JSON) file with the following format:
\begin{enumerate}
\item \textbf{image\_name} (str): full name of MR image file with the data format extension
\item \textbf{crop} (str): plant species of the MR image
\item  \textbf{has\_root} (int): flag for whether image has roots (0 does not contain roots, 1 contains roots)
\item  \textbf{binary\_mask} (str): variable indicating whether the image has a corresponding pixel-level root segmentation mask. If so, the value is the full name of the corresponding pixel-level segmentation mask. Otherwise, the value is `N/A'.
\item  \textbf{location} (str): abbreviation of the location where the image was collected
\item  \textbf{tube\_num} (str): MR tube number where the image was collected
\item \textbf{date} (str): date (Year-Month-Day) when the image was collected
\item  \textbf{depth} (str):  depth at which the was image collected
\item  \textbf{dpi} (str):  DPI of the collected image\\

In addition to the above, switchgrass MR annotations also contain the following: 

\item \textbf{row\_idx} (str): the row index of the image in the original spanning image
\item \textbf{col\_idx} (str): the column index of the image in the original spanning image
\end{enumerate}

\subsection{Pixel-level Annotation for Switchgrass}
\begin{figure*}[t]
\begin{center}
    \subfloat[Spanning Switchgrass Image]{%
      \fbox{\includegraphics[width=0.3\linewidth]{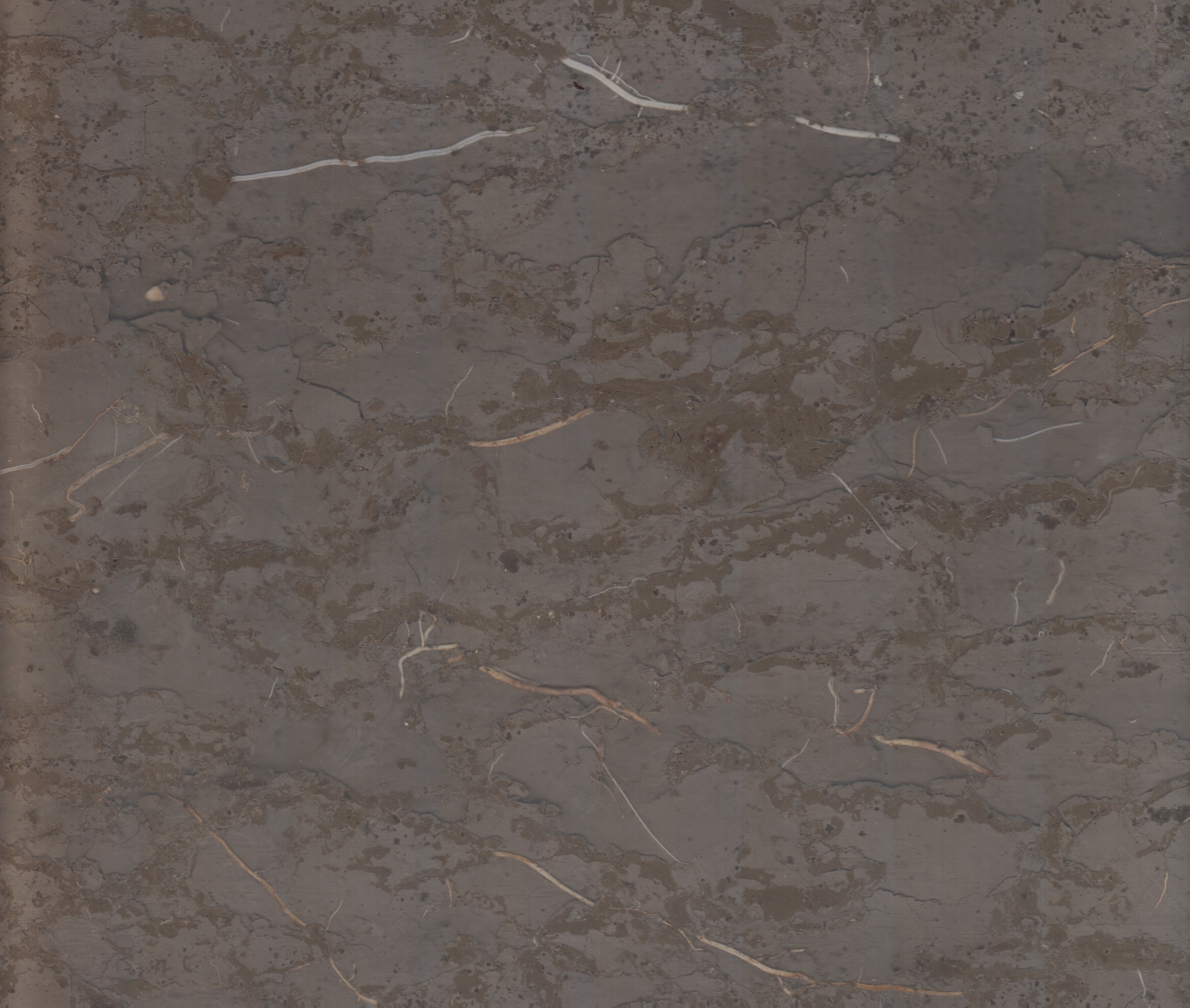}}}
    \subfloat[Manually Annotated Mask]{%
      \fbox{\includegraphics[width=0.3\linewidth]{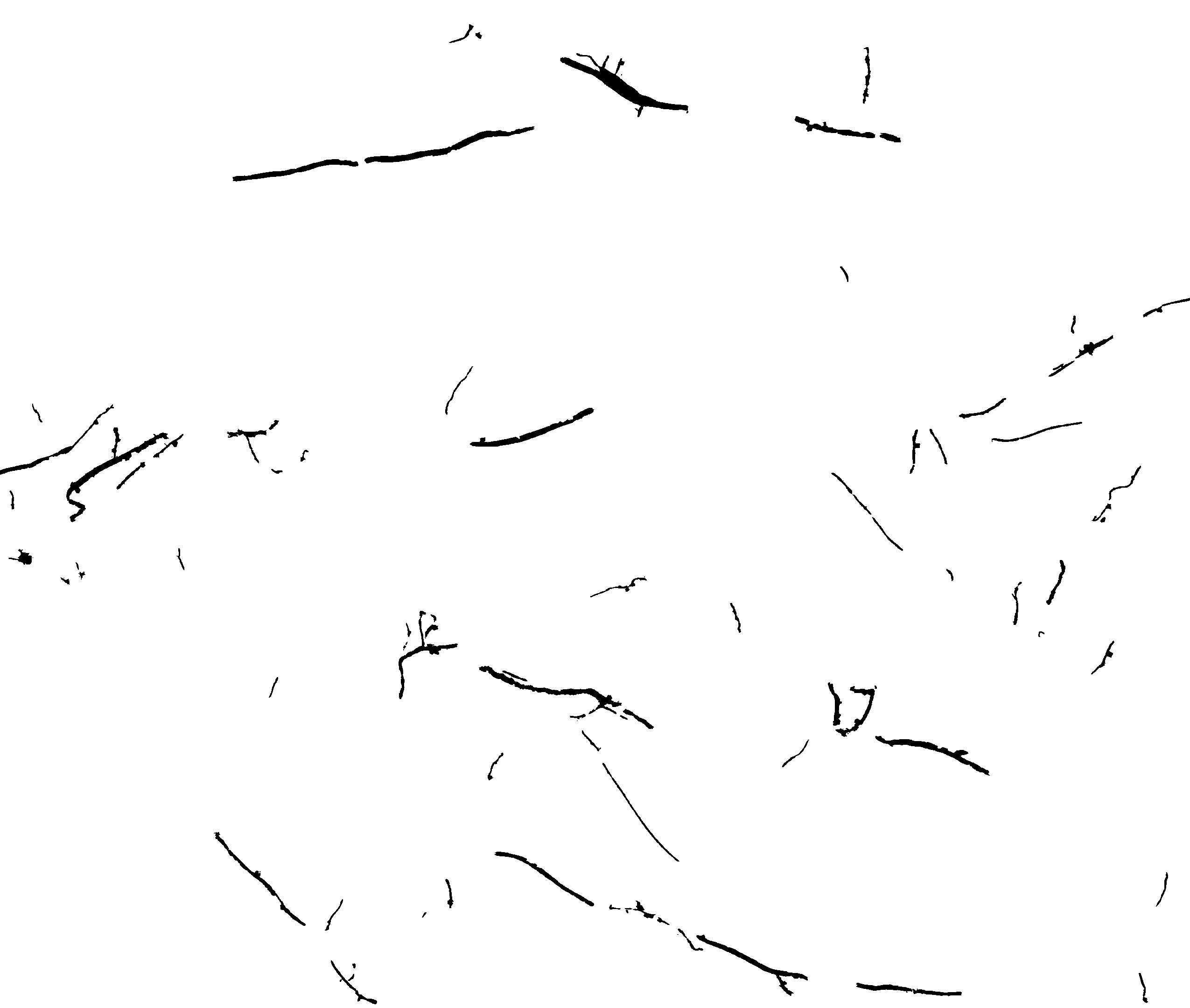}}} \\
      
    \subfloat[Spanning Switchgrass Image]{%
      \fbox{\includegraphics[width=0.3\linewidth]{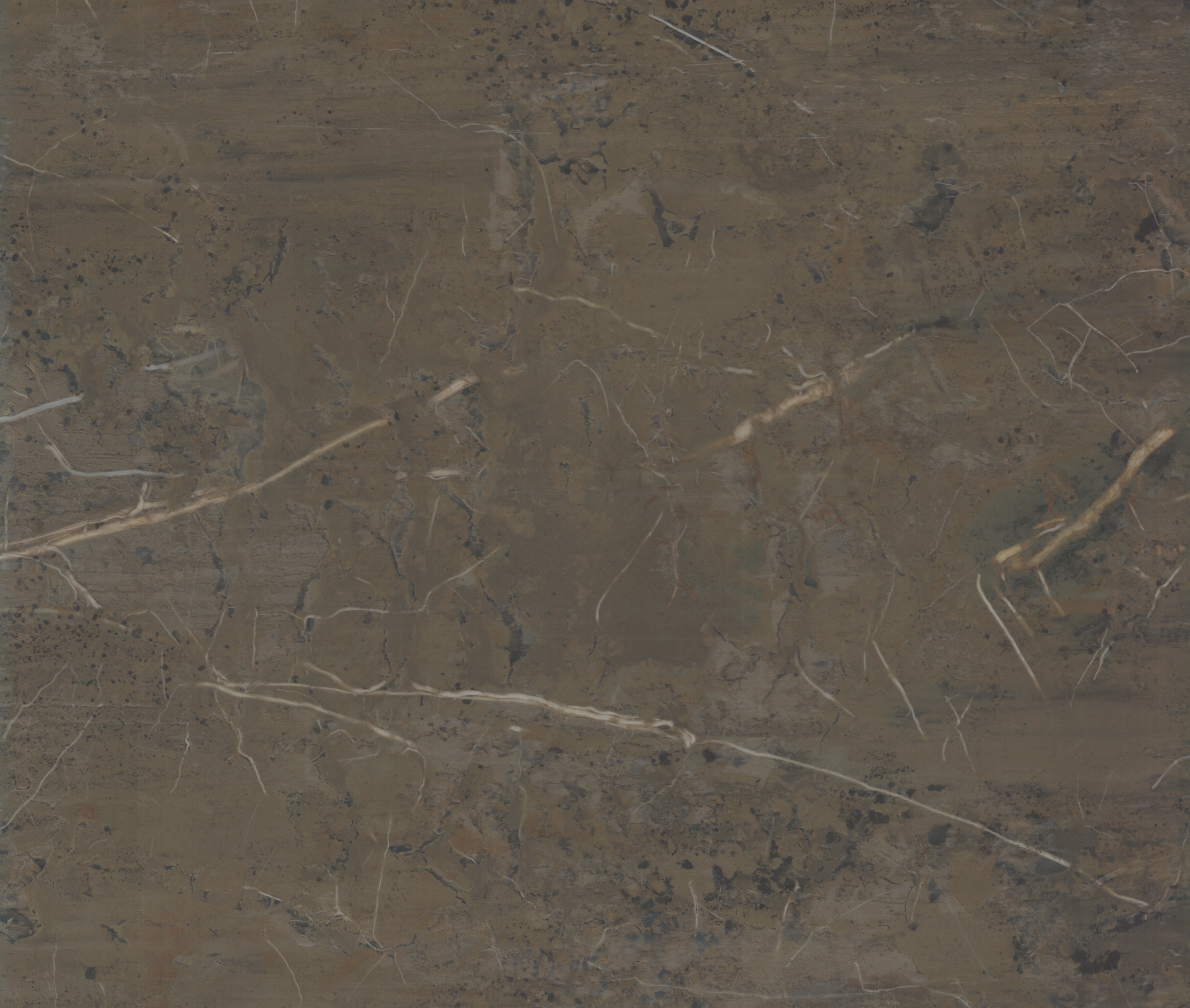}}}
    \subfloat[U-net Predicted Mask]{%
      \fbox{\includegraphics[width=0.3\linewidth]{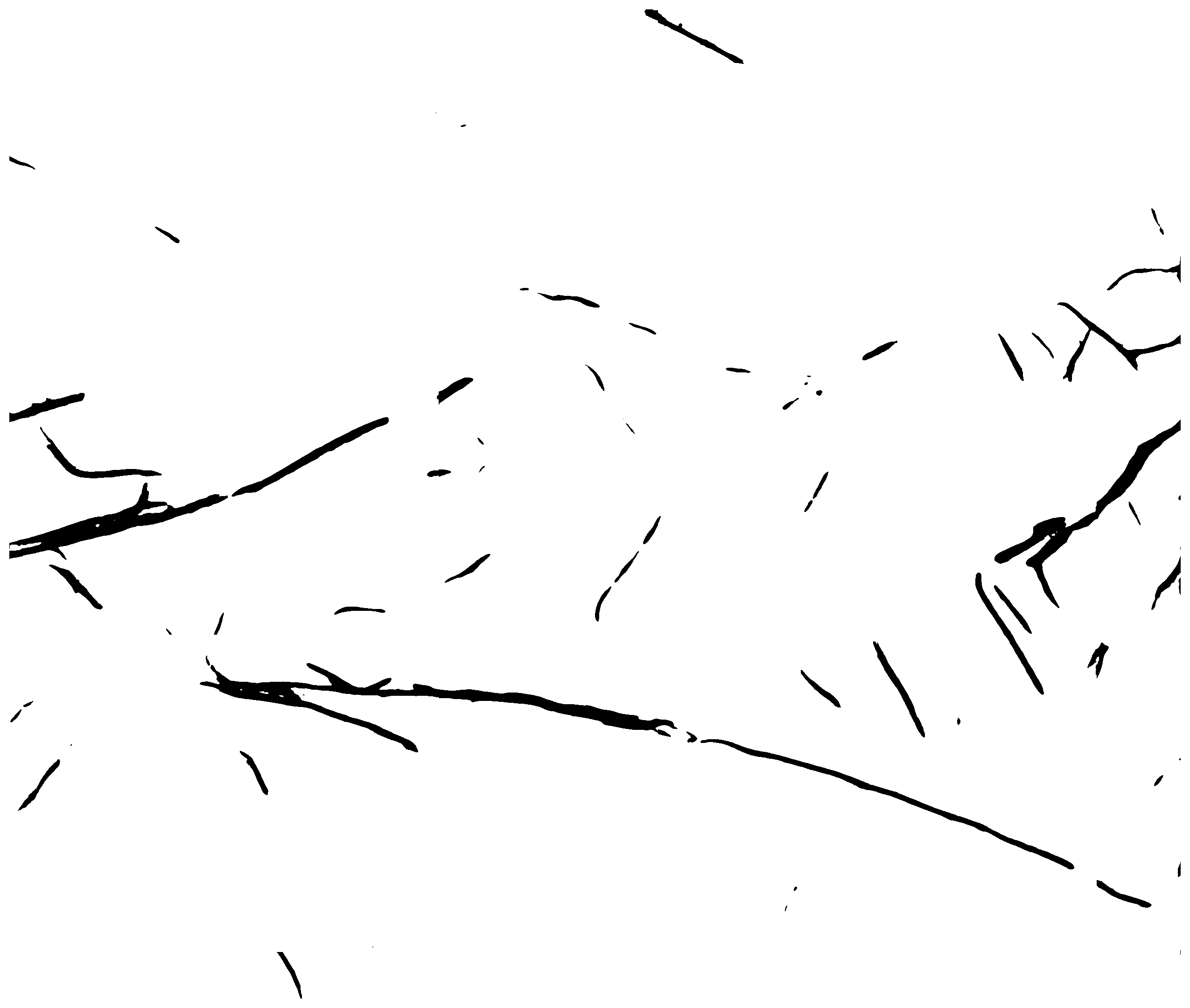}}}
    \subfloat[Technician Refined Mask]{%
      \fbox{\includegraphics[width=0.3\linewidth]{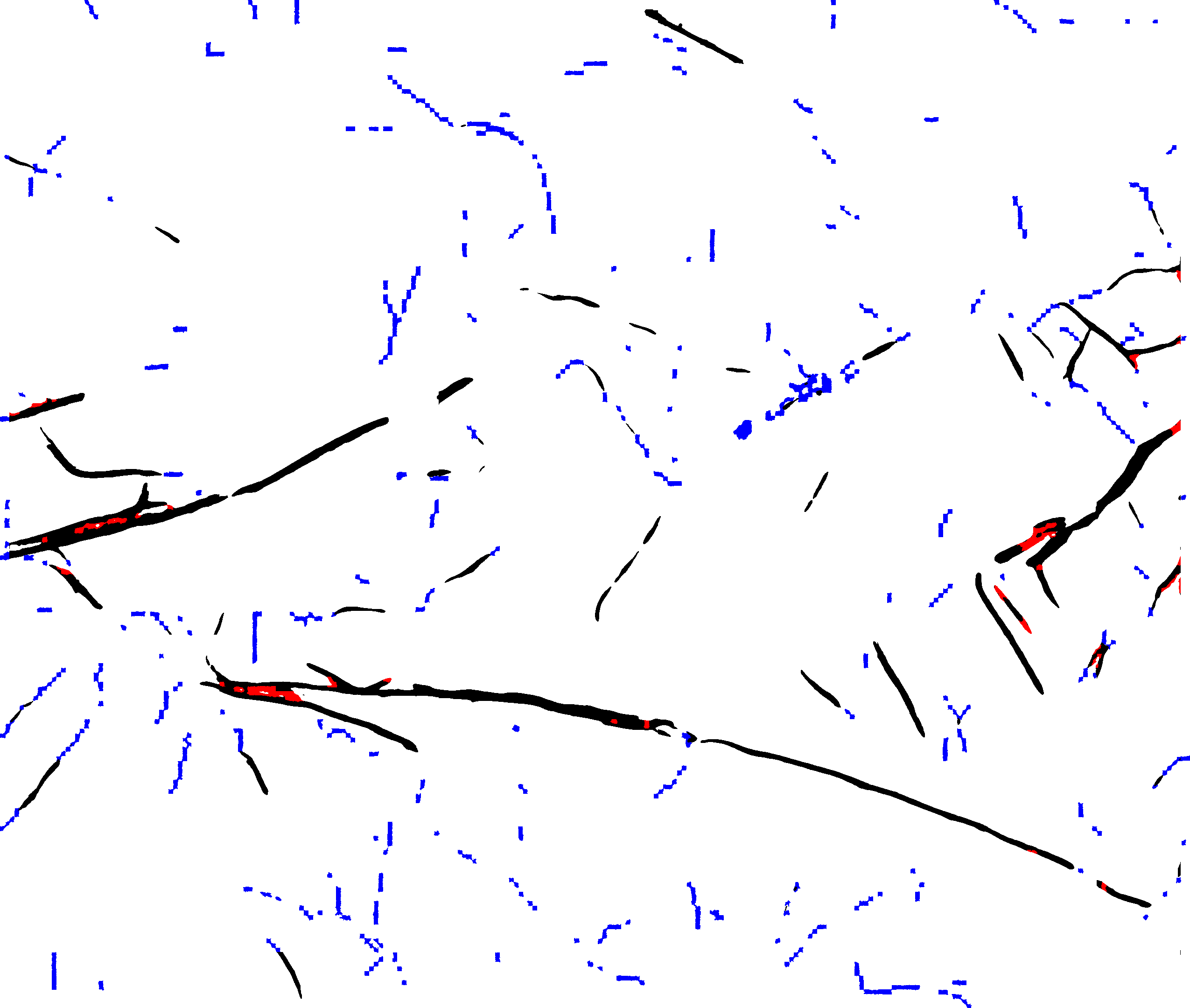}}}      
     
\end{center}
  \caption{Examples switchgrass root MR images and corresponding annotated masks: (a) witchgrass MR image, (b) manually annotated ground truth masks, (c) switchgrass MR image, (d) predicted segmentation masks generated by pre-trained U-net, and (e) technician refined U-net-predicted mask. In (c), black roots are unchanged, blue roots were added by the technician, and red roots were deleted.}
  \label{fig:Image&GT_SG} 
\end{figure*}

The switchgrass dataset can be divided into three components: (a) 8625 images only paired with image-level annotation (no pixel-level annotation masks); (b) 600 images paired with image-level and pixel-level annotation generated manually by technicians; and (c) 3312 images paired with image-level and pixel-level annotation which were generated by technicians refining the segmentation results of a pre-trained U-net \cite{ronneberger2015u}. The annotation methods for (b) and (c) are described below:

\textbf{(b) Manual annotation by technicians for switchgrass MR images}: The raw switchgrass MR spanning images each have a size of $2550 \times 2160$.  Given the fine, narrow switchgrass roots and the size of these images, it is extremely time-consuming to manually annotate each pixel. To improve the efficiency of the annotation process, we ran the Simple Linear Iterative Clustering (SLIC) algorithm \cite{achanta2010slic} on raw images to segment images into superpixels by clustering spatially-contiguous pixels into groups according to color. Then, the technicians annotated the imagery on a superpixel level such that all the pixels belonging to the same superpixel share the same label. The fine roots are challenging to accurately delineate and over-segment using SLIC due to their small size and, often, similarity in color to the background. Thus, after annotating images on the superpixel level, the technicians refined the masks by manually fine-tuned the labels on a pixel-level for missing or inaccurate roots and root edges. This dataset contains 600 switchgrass MR root images annotated using this method. Examples of these switchgrass images and the corresponding manually annotated masks are shown in Figure \ref{fig:Image&GT_SG} (a) and (b), respectively.

\textbf{(c) AI prediction aided annotation for switchgrass MR images:}
To further improve the annotation efficiency, we also explored the possibility of using the prediction results of a pre-trained U-net. Specifically, we used a U-net model pre-trained on a large peanut root dataset and fine-tuned on a limited switchgrass root dataset \cite{xu2020overcoming} to generate predicted segmentation masks for cropped switchgrass images. Then, technicians refined the predicted segmentation masks by manually adding or deleting the roots on the pixel-level. There are 3312 switchgrass root MR images annotated using this method. Examples of a spanning switchgrass image, the predicted segmentation mask generated by U-net, and technician refined U-net-predicted mask are shown in Figure \ref{fig:Image&GT_SG} (c), (d) and (e), respectively. 

\section{Appendix C: Intended Use of the Dataset} \label{sec:use}
\textbf{Supervised root segmentation}. Semantic segmentation of plant roots from the background is the first step before any further MR analysis. Generally, training images with pixel-level annotation are required for this task. Several models (e.g., fully convolutional networks\citep{long2015fully}, SegNet\citep{badrinarayanan2015segnet}, U-net\citep{ronneberger2015u}, and DeepLab\citep{chen2018deeplab}) have achieved success in segmentation of plant roots and leaves\citep{chen2018automatic,zhu2018data,yu2020weakly,yu2020root}. With this dataset, we expect to boost the research of more advanced models and algorithms for plant root segmentation applications given pixel-level labeled training data.

\textbf{Weakly-supervised root segmentation}. Generating pixel-level ground truth masks for MR imagery is incredibly time-consuming and tedious.  Thus, effort in weakly-supervised semantic segmentation in which only image-level labels are used to train a model would greatly boost MR segmentation efforts. Instead of requiring a significant effort of labeling MR imagery for every new field and plant species (with varying soil and root properties), images (or sub-images) could be much more efficiently labeled with image-level labels.  

\textbf{Segmentation across location and species}. Several MR segmentation approaches have been developed in the literature. However, each of the approaches needs to be re-trained to achieve adequate performance on any new plant species or new location with different soil properties.  One use of this dataset is to develop approaches that can be effectively and efficiently transferred to new locations and species.  

\textbf{Root feature analysis}. For each species, we collected images at a different depth, location over a period of time to capture the growth process of the plant root system. The collection location, depth, and time information are saved in the image-level label for each image. Root features such as root color, root thickness, root length, root surface area, and the number of fine roots, etc. can be extracted from root imagery. Given a specific location and depth, the changes of young roots to old roots were recorded over time. This dataset can be used in the time-series analysis of these root features, which is valuable to identify the relationship between root age and root features. The depth information plays an important role in terms of the structure of the whole root system. This dataset can also be used for root alignment applications to reconstruct the root system by combining imagery from a different depth. Finally, this dataset covers $6$ different species. Comparing and characterizing the root system across species can also be studied using this dataset.

\section{Appendix D: Training Setup}
We used U-net based model \cite{xu2020overcoming} and IRNet \cite{ahn2019weakly} to benchmark the dataset for supervised learning and weakly-supervised learning, respectively. The architecture of U-net model has five convolutional blocks in encoder and four trans-convolutional blocks in decoder. The details of architecture can be found in \cite{xu2020overcoming}.  Adamax optimizer was used with learning rate 1e-4 and weight decay 5e-4. Binary cross-entropy loss was used to calculate loss between predicted segmentation masks and ground truth masks. The models are trained on suggested training set for 300 epochs. For weakly-supervised learning using IRNet, we first trained the CAM step (ResNet50 backbone) on training set using image-level label to classify the root images vs. non-root images. We followed the training strategies in \cite{ahn2019weakly}. The pseudo ground truth masks are generated for both training set and validation set by thresholding (0.9 as shown in \citep{yu2020weakly}) the attention map. Then, the U-net model were training on these pseudo binary masks using the same training setup as the supervised learning. The model with the smallest error on the validation set will be used to evaluate the segmentation performance on test set.

\begin{figure*}[t]
\begin{center}
    \subfloat[Cotton]{%
      \includegraphics[width=0.192\linewidth]{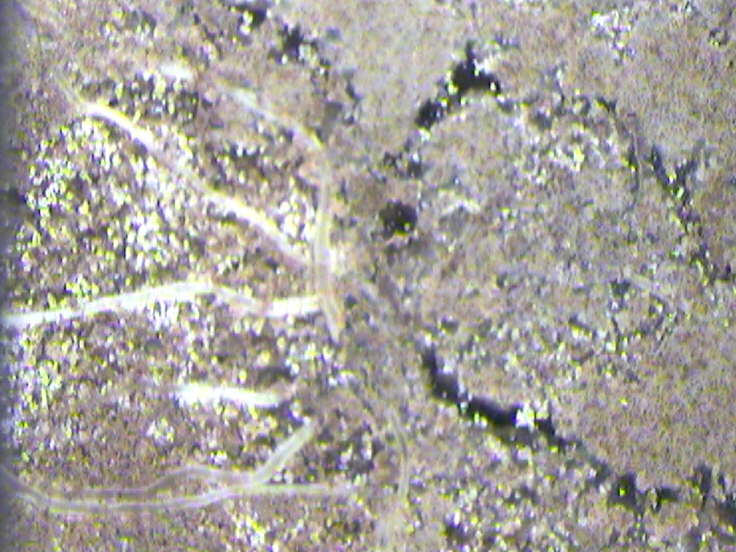}} \hfill
    \subfloat[Papaya]{%
      \includegraphics[width=0.192\linewidth]{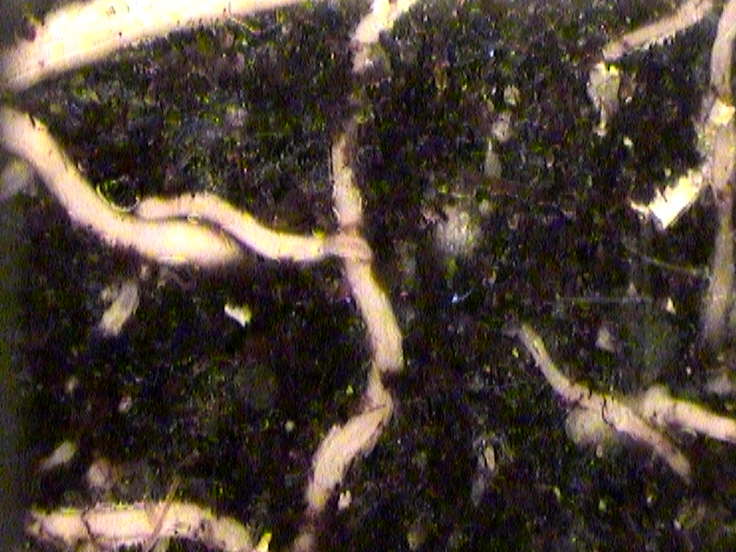}} \hfill
    \subfloat[Peanut]{%
      \includegraphics[width=0.192\linewidth]{./2016SHELTERS_T001_L041_2016.06.07_140406_001_BAZ.jpg}} \hfill
    \subfloat[Sesame]{%
      \includegraphics[width=0.192\linewidth]{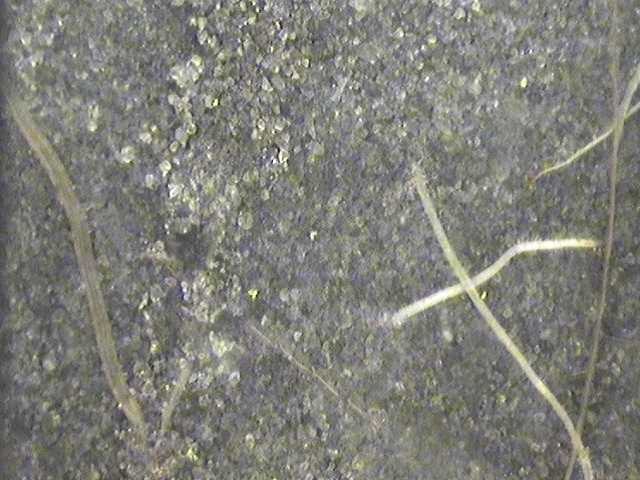}} \hfill
    \subfloat[Sunflower]{%
      \includegraphics[width=0.193\linewidth]{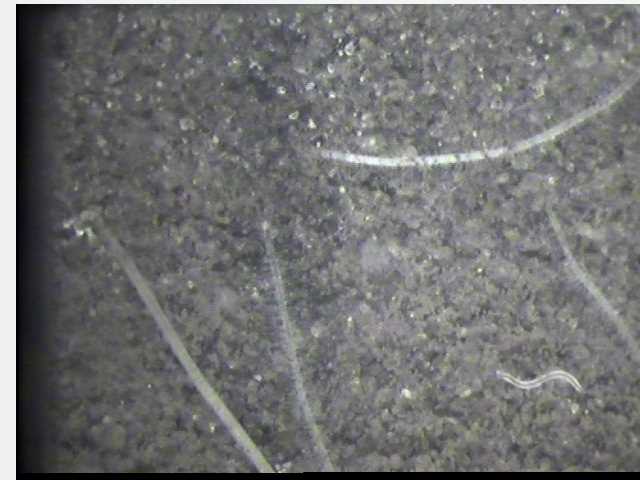}} \hfill
    \\
  \subfloat{%
        \fbox{\includegraphics[width=0.18\linewidth]{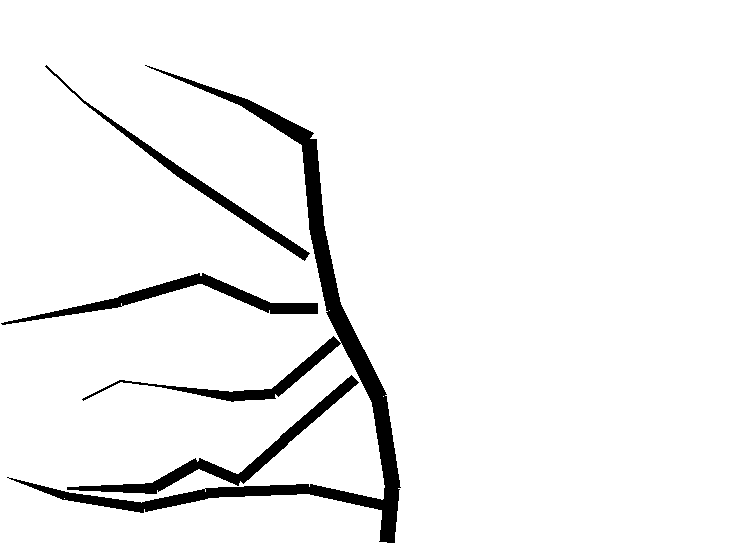}}} \hfill
  \subfloat{%
        \fbox{\includegraphics[width=0.18\linewidth]{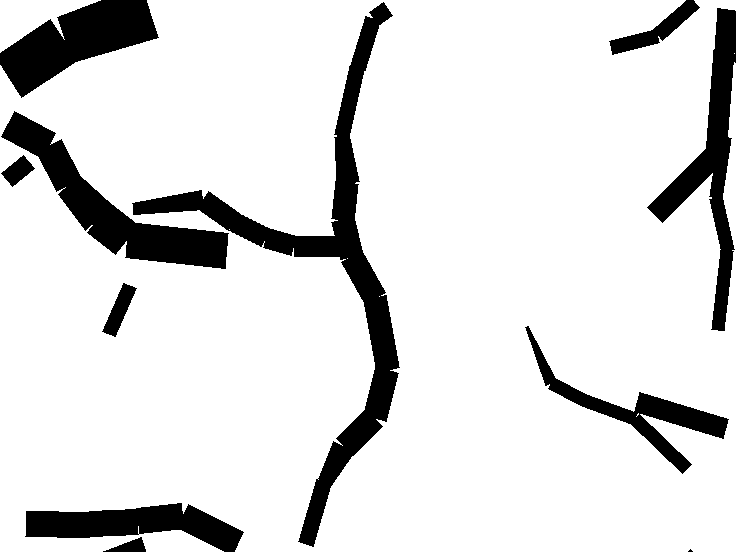}}} \hfill
  \subfloat{%
        \fbox{\includegraphics[width=0.18\linewidth]{./GT_2016SHELTERS_T001_L041_2016.06.07_140406_001_BAZ.png}}} \hfill
  \subfloat{%
        \fbox{\includegraphics[width=0.18\linewidth]{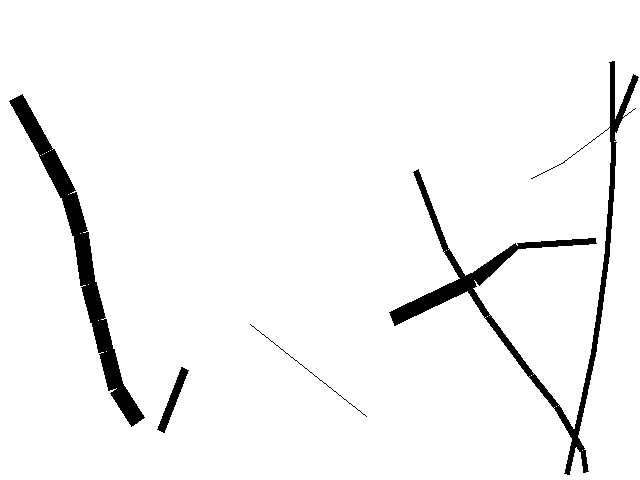}}} \hfill
  \subfloat{%
        \fbox{\includegraphics[width=0.18\linewidth]{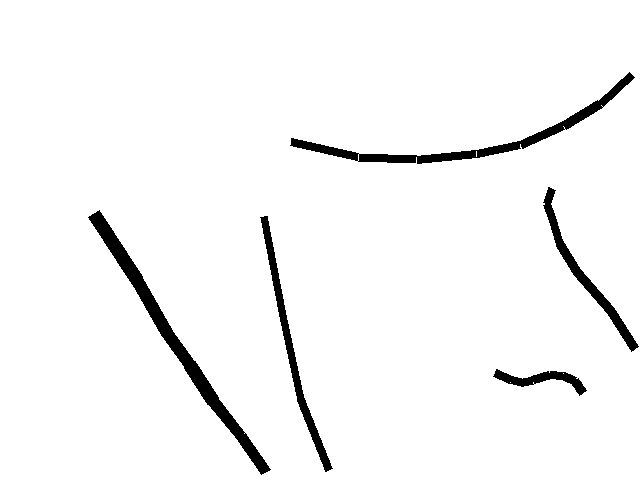}}} \hfill       
\end{center}
  \caption{Examples of selected raw root MR images (top row) and manually annotated ground truth masks (bot row) using WinRHIZO Tron software.}
  \label{fig:Image&GT} 
\end{figure*}

\end{document}